\documentclass[12pt]{article}
\pdfoutput=1
\usepackage[utf8]{inputenc}
\usepackage[T1]{fontenc}
\usepackage{float}
\usepackage{authblk}
\usepackage[normalem]{ulem}
\usepackage{xcolor}
\usepackage{amsmath}
\usepackage{amsthm}
\usepackage{adjustbox}
\usepackage{graphicx}
\usepackage{mathtools}
\usepackage{amsfonts}
\usepackage{fullpage}
\usepackage{subcaption}
\usepackage[style=numeric,sorting=none,maxnames=99]{biblatex}
\usepackage{listings}
\usepackage{xcolor}

\definecolor{codegreen}{rgb}{0,0.6,0}
\definecolor{codegray}{rgb}{0.5,0.5,0.5}
\definecolor{codepurple}{rgb}{0.58,0,0.82}
\definecolor{backcolour}{rgb}{0.95,0.95,0.92}

\lstdefinestyle{mystyle}{
    backgroundcolor=\color{backcolour},   
    commentstyle=\color{codegreen},
    keywordstyle=\color{magenta},
    numberstyle=\tiny\color{codegray},
    stringstyle=\color{codepurple},
    basicstyle=\ttfamily\footnotesize,
    breakatwhitespace=false,         
    breaklines=true,                 
    captionpos=b,                    
    keepspaces=true,                 
    numbers=left,                    
    numbersep=5pt,                  
    showspaces=false,                
    showstringspaces=false,
    showtabs=false,                  
    tabsize=2
}

\usepackage[colorlinks = true,
            linkcolor = teal,
            urlcolor  = teal,
            citecolor = teal]{hyperref}

\title{\texttt{pymdp}: A Python library for active inference\\in discrete state spaces}

\author[1,2,3,4]{Conor Heins\thanks{Correspondence: \href{mailto:conor.heins@gmail.com}{conor.heins@gmail.com}; \href{mailto:tschantz.alec@gmail.com}{tschantz.alec@gmail.com}}}
\author[4,5]{Beren Millidge}
\author[6]{Daphne Demekas}
\author[4,7,8]{Brennan Klein}
\author[9]{Karl Friston}
\author[1,2,3]{Iain D. Couzin}
\author[4,10,11]{Alexander Tschantz$^*$}

\affil[1]{Department of Collective Behaviour, Max Planck Institute of Animal Behavior,\protect\\78457 Konstanz, Germany}
\affil[2]{Centre for the Advanced Study of Collective Behaviour, 78457 Konstanz, Germany}
\affil[3]{Department of Biology, University of Konstanz, 78457 Konstanz, Germany}
\affil[4]{VERSES Research Lab, Los Angeles, California, USA}
\affil[5]{MRC Brain Networks Dynamics Unit, University of Oxford, Oxford, UK}
\affil[6]{Department of Computing, Imperial College London, London, UK}
\affil[7]{Network Science Institute, Northeastern University, Boston, MA, USA}
\affil[8]{Laboratory for the Modeling of Biological and Socio-Technical Systems,\protect\\Northeastern University, Boston, USA}
\affil[9]{Wellcome Centre for Human Neuroimaging, Queen Square Institute of Neurology,\protect\\University College London, London WC1N 3AR, UK}
\affil[10]{Sussex AI Group, Department of Informatics, University of Sussex, Brighton, UK}
\affil[11]{Sackler Centre for Consciousness Science, University of Sussex, Brighton, UK}

\begin{document}
\maketitle
\pagenumbering{arabic}

\section*{\label{sec:introduction}Statement of need}

Active inference is an account of cognition and behavior in complex systems which brings together action, perception, and learning under the theoretical mantle of Bayesian inference \cite{friston_reinforcement_2009, friston_active_2012, friston2015active, friston2017process}. Active inference has seen growing applications in academic research, especially in fields that seek to model human or animal behavior \cite{parr2020prefrontal, holmes2021active, adams2021everything}. The majority of applications have focused on cognitive neuroscience, with a particular focus on modelling decision-making under uncertainty. Nonetheless, the framework has broad applicability and has recently been applied to diverse disciplines, ranging from computational models of psychopathology \cite{montague2012computational, schwartenbeck2015dopaminergic, smith2020imprecise, smith2021greater}, control theory \cite{baltieri2019pid, millidge2020relationship, baioumy2021towards} and reinforcement learning  \cite{tschantz2020reinforcement, tschantz2020scaling, sajid2021active, fountas2020deep,millidge2020deep}, through to social cognition \cite{adams2021everything, hipolito2021enactive, wirkuttis2021leading, tison2021communication} and even real-world engineering problems \cite{martinez2021probabilistic, moreno2021pid, fox2021active}. While in recent years, some of the code arising from the active inference literature has been written in open source languages like Python and Julia  \cite{ueltzhoffer2018deep, van2019simulating, tschantz2020learning, ccatal2020learning, millidge2020deep}, to-date, the most popular software for simulating active inference agents is the \texttt{DEM} toolbox of \texttt{SPM} \cite{friston2008variational, smith2022step}. \texttt{SPM} is a MATLAB library originally developed for the statistical analysis and modelling of neuroimaging data, such as data collected by functional magnetic resonance imaging (fMRI) or magneto- and electro-encephalographic (MEG/EEG) \cite{penny2011statistical} methods. The \texttt{DEM} toolbox, a sub-library of \texttt{SPM}, was originally developed to simulate and perform Bayesian estimation of dynamical systems \cite{friston2008variational}, but in the last decade it has been augmented with a series of demonstrative scripts and simulation routines related to active inference and the Free Energy Principle more broadly \cite{friston2010free,friston2013life, friston2019free}.

Simulations of active inference are commonly performed in discrete time and space \cite{da2020active, friston2015active}. This is partially motivated by the mathematical tractability of performing inference with discrete probability distributions, but also by the intuition of modelling choice behavior as a sequence of discrete, mutually-exclusive choices, in e.g. psychophysics or decision-making experiments. The most popular generative models -- used to realize active inference in this context -- are partially-observable Markov Decision Processes or \textit{POMDPs} \cite{kaelbling1998planning}. POMDPs are state-space models that model the environment in terms of hidden states that stochastically change over time, as a function of both the current state of the environment as well as the behavioral output of an agent (control states or actions). Crucially, the environment is \textit{partially-observable}, i.e. the hidden states are not directly observed by the agent, but can only be inferred through observations that relate to hidden states in a probabilistic manner, such that observations are modelled as being generated stochastically from the current hidden state.

In most POMDP problems, an agent is tasked with both inferring the hidden states and selecting a sequence of control states or actions to change the hidden states in a way that leads to desired outcomes (maximizing reward, or occupancy within some preferred set of states). \texttt{DEM} contains a reliable, reproducible set of functions for simulating active inference agents equipped with such generative models: they include \texttt{spm\_MDP\_VB\_X.m}, \texttt{spm\_MDP\_game.m}, and -- recently introduced for simulating `sophisticated' active inference -- \texttt{spm\_MDP\_VB\_XX.m} \cite{friston2021sophisticated}. Despite its robustness and widespread use among active inference researchers, \texttt{spm\_MDP\_VB\_X.m} is a single function, meaning that any active inference simulation using the function has to comply with the constraints implied by its structure and control flow. Although options can be specified that initiate particular sub-routines or variants of active inference, it is still not straightforward to construct a custom active inference process from scratch. In practice, this means that novel, bespoke applications require researchers to manually adapt parts of \texttt{spm\_MDP\_VB\_X.m} for their own purposes, which limits the general reproducibility and adaptability of academic active inference research, especially for new practitioners. In addition, since all of \texttt{DEM} is written in MATLAB, using the toolbox can be prohibitive due to the cost of a MATLAB license, especially for researchers who are unaffiliated with institutions. Increasing interest in active inference, manifested both in terms of sheer number as well as diversifying applications across scientific disciplines, has thus created a need for generic, widely-available, and user-friendly code for simulating active inference in open-source scientific computing languages like Python. The software we present here, \texttt{pymdp}, represents a significant step in this direction: namely, we provide the first open-source package for simulating active inference with discrete state-space generative models. The name \texttt{pymdp} derives from the fact that the package is written in the \textbf{Py}thon programming language and concerns discrete, Markovian generative models of decision-making, which take the form of Markov Decision Processes or \textbf{MDP}s.

We developed \texttt{pymdp} to increase the accessibility and exposure of the active inference framework to researchers, engineers, and developers with diverse disciplinary backgrounds. In the spirit of open-source software, we also hope that it spurs new innovation, development, and collaboration in the growing active inference community.

\section*{\label{sec:summary}Summary}

\texttt{pymdp} offers a suite of robust, tested, and modular routines for simulating active inference agents equipped with \emph{partially observable Markov Decision Process} (POMDP) generative models. Mathematically,  a POMDP comprises a joint distribution over observations $o$, hidden states $s$, control states $u$ and hyperparameters $\phi$: $P(o, s, u, \phi)$. This joint distribution further factorizes into a set of categorical and Dirichlet distributions: the likelihoods and priors of the generative model. With \texttt{pymdp}, one can build a generative model using a set of prior and likelihood distributions, initialize an agent, and then link it to an external environment to run active inference processes - all in a few lines of code. The agent and environment API is built according to the standardized framework of OpenAIGym commonly used in reinforcement learning, where an \texttt{Agent} and \texttt{Environment} class recursively exchange observations and actions over time \cite{brockman2016openai}.

In order to enhance the user-friendliness of \texttt{pymdp} without sacrificing flexibility, we have built the library to be highly modular and customizable, such that agents in \texttt{pymdp} can be specified at a variety of levels of abstraction with desired parameterizations. In the next section, we provide an overview of the structure of the package.

\subsection*{\label{subsec:package_structure}Package structure}

\subsubsection*{\label{subsubsec:agent}The \texttt{Agent} Class}

The high-level API offered by \texttt{pymdp} is the \texttt{Agent} class. Instantiating an \texttt{Agent} allows the user to abstract away the various optimization routines and sub-operations that make up an active inference process, e.g. state estimation, action selection, and learning. The various sub-routines of active inference are themselves abstracted as user-friendly methods of \texttt{Agent} (such as \texttt{self.infer\_states(obs)}), calls to which will run the corresponding function. 

\subsubsection*{\label{subsec:modules}Modules}

The methods of \texttt{Agent} themselves call functions from different sub-modules of \texttt{pymdp}. These submodules can be roughly divided into three sorts of operations: \textit{perception}, \textit{action} and \textit{learning}. An attractive feature of active inference is that various cognitive processes naturally emerge as different variants of Bayesian inference. For instance, instantaneous inference about dynamically-changing hidden states is often analogized to  \textit{perception} (c.f. perception as inference \cite{von1910treatise,gregory1980perceptions,hinton1983optimal, dayan1995helmholtz}), whereas inference about slower-changing variables (statistical regularities in the environment) is analogized to learning \cite{friston2016active}. Moreover, \textit{action} is treated as a process of inference, where agents select actions by inferring a distribution over control states or sequence of control states \cite{friston2017process}. Each module of \texttt{pymdp} thus performs inference with respect to different components of a POMDP generative model -- we summarize them briefly below. 

The \texttt{inference} library of \texttt{pymdp} contains a set of functions for performing hidden state inference or state estimation. These are the core functions that allow agents to update their beliefs about the discrete hidden state of the environment, given observations. Functions from this library are called by the \texttt{self.infer\_states()} method of \texttt{Agent}. Specific arguments can be passed into the \texttt{Agent} constructor to specify the type and parameterization of the algorithm used to perform hidden state inference.

The \texttt{control} library of \texttt{pymdp} contains functions for inferring policies and sampling actions from the posterior beliefs about control states.\footnote{In active inference `control states' refer to the random variables in the generative model and approximate posterior, and can be thought of as the agent's representation of its actions. Actions themselves are \emph{realizations} of these random variables, sampled from the posterior over control states.} These functions are called internally by the \texttt{self.infer\_policies()} method of \texttt{Agent}.

Finally, the \texttt{learning} library of \texttt{pymdp} contains the functions necessary for the agent to update hyperparameters of its generative model, i.e. Dirichlet parameters over the categorical prior and likelihood distributions. These functions are called internally by methods like \texttt{self.update\_A()}, \texttt{self.update\_B()}, and \texttt{self.update\_D()} of \texttt{Agent}.

For a more detailed overview of the functionality offered by each of \texttt{pymdp}'s modules, please see \nameref{sec:Appendix_B}.

\begin{lstlisting}[language=Python, caption=Minimal example of running active inference with \texttt{pymdp}., label={lst:usage1}]   
    from pymdp.agent import Agent
    
    # here you would set up your generative model 
    my A = ...
    my_B = ...
    
    # instantiate your agent with a call to the Agent() constructor
    my_agent = Agent(A=my_A, B=my_B, C=my_C, D=my_D)
    
    # define an environment
    my_env = Env()
    
    # set an initial action
    action = initial_action
    
    for t in range(T):
        o_t = my_env.step(action)
        my_agent.infer_states(o_t)
        my_agent.infer_policies() 
        action = my_agent.sample_action() 
    
\end{lstlisting}

\section*{\label{sec:usage}Usage}

Specifying a generative model is central to active inference, and to Bayesian modelling in general. Intuitively, a generative model is a probabilistic specification of how data or sensory observations are generated. In the context of Bayesian agent-based models, the generative model represents an agent's probabilistic internal model of its environment, comprising a set of structural assumptions about how the world generates observations and how action changes the world. In discrete-state and -time active inference models, we typically assume the generative model is a \emph{partially observable Markov Decision Process} or POMDP, comprising \emph{observations} the agent receives, \emph{hidden states} of the world, and \emph{actions} the agent can take to influence hidden states. Hidden states are called `hidden' precisely because the agent can never directly access them, but can only \emph{infer} them via observations. The POMDP structure assumes that at each timestep, the observation is generated by the current hidden state, while the hidden state itself changes over time as a function of its current setting and some control state (i.e. action).
Mathematically, a generative model is usually expressed as a joint probability distribution $P(o,s,u, \phi)$ over observations $o$, hidden states $s$, control states $u$, and parameters $\phi$. This joint distribution is called a \emph{generative} model because it can be used to sample (or generate) sequences of potential observations according to the probabilistic structure encoded in the model.

Specifying the generative model in terms of discrete probability distributions is the first step to building an active inference agent in
\texttt{pymdp}. Below, we overview the steps involved in building a generative model to provide intuition and illustrate its simplicity in the special case of POMDPs.

\subsection*{\label{subsec:pomdp_genmodel}The POMDP generative model}

The POMDP generative model assumed by \texttt{pymdp} is a discrete-time and space generative model that, like any probability distribution, can be factorized into a product of conditional distributions (likelihoods) and marginals (priors). The most important of these distributions -- when writing down a generative model in \texttt{pymdp} -- are 1) the observation likelihood $P(o_\tau|s_\tau)$, which represents the agent's beliefs about how hidden states $s$ generate observations $o$ and 2) the transition model $P(s_{\tau} | s_{\tau}, u_{\tau-1})$, which represents the agent's beliefs about how hidden states at some time $\tau-1$ cause hidden states at the next time $\tau$, conditioned on some control states (actions) $u_{\tau-1}$. The agent also has a prior over initial hidden states $P(s_1)$, which represents the agent's baseline belief, before gathering any observations, about the probability of the different hidden states at the first timestep. For the sake of mathematical convenience, we describe POMDP generative models with a finite time horizon $T$, but note that in general \texttt{pymdp} does not require a finite time horizon, so active inference agents can be theoretically run indefinitely (e.g. in streaming applications). Finally, there is an additional prior distribution over observations, $P(o_{1:T})$, which specifies an agent's goals as a desired distribution over observations. 
In active inference, goals and desires are encoded as a prior in the generative model, such that the probability that the model assigns to some configuration of observations, hidden states, control states and prior parameters is only maximized when sampling preferred observations (e.g. if my model assigns high probability to observing body temperature to be 37 degrees, prior preferences are realized when these observations are sampled). As we will see in the following sections, active inference suggests that perception, action and learning all work to maximize the marginal likelihood of observations. Prior preferences only come into play during policy selection because only unobserved outcomes in the future are random variables (see the section on \nameref{sec:control} in Appendix B for details).

The local dependence in time, captured by one-step conditional dependence between hidden states in the transition likelihood, is what renders POMDPs \textit{Markovian} generative models. As such, a general expression for the joint distribution over $o$, $s$, $\pi$ and $\phi$ is as follows:
\begin{align}
    P(o_{1:T}, s_{1:T}, \pi, \phi) = P(\phi)P(s_1)P(\pi)\prod_{\tau = 2}^{T}P(s_{\tau} | s_{\tau-1}, \pi; \phi)\prod_{\tau = 1}^{T} P(o_\tau | s_\tau; \phi)  \label{eq:POMDP_gen_model}
\end{align}

We have replaced $u$ here with $\pi$ to represent policies, or sequences of control states $u$, i.e. $\pi = \{u_i, u_j, u_k, ...\}$. Control states can be formally related to policies by writing down an additional likelihood, a `policy-to-control' mapping $P(u_{\tau} |\pi)$, that links a given policy to the control state it entails at time $\tau$. Under active inference, agents also perform \emph{inference about policies}, which naturally entails goal-directed and uncertainty-resolving behavior (see the section \nameref{sec:EFE} in Appendix B for details on policy inference). Finally, we capture any additional parameters as a single vector of hyperparameters $\phi$, which might correspond to the parameters of Dirichlet priors over the likelihood distributions $P(o_\tau|s_\tau)$ and $P(s_{\tau}|s_{\tau-1}, \pi)$, for instance. In active inference, inference about these hyperparameters is often assumed to occur on a slower timescale than inference about hidden states and policies: therefore, this process is referred to in the active inference literature as learning \cite{friston2016active, schwartenbeck2019computational} (see \nameref{sec:learning} for more details on hyperparameter inference).

Equipped with a generative model, active inference (and Bayesian inference more generally) entails inference over latent variables $s$, $\pi$ and $\phi$, given some observations $o_t$ gathered over time. For a description of the mathematical basis of this inference -- and how it is implemented algorithmically in \texttt{pymdp} -- please refer to \nameref{sec:Appendix_B}. Now that we have expressed the POMDP generative model formally, highlighting its important components for active inference, we can move on to the representation of these distributions in \texttt{pymdp}.

\subsection*{\label{subsec:building_blocks}Building blocks}

\texttt{pymdp} considers generative models of discrete states that evolve in discrete time. This means that there are an integer number of discrete levels of both the states of the environment and of the observations. Because of this fundamental discreteness, a natural way to represent the distributions of the generative model is by using \emph{categorical} distributions, which assign a probability value between $0$ and $1$ to each discrete outcome level of the distribution's sample space, with the usual constraint that the sum of the probabilities over levels is $1$. Mathematically, we refer to categorical distributions with the notation $P(x) = Cat(\boldsymbol{\phi}), \boldsymbol{\phi} \in \{z \in \mathbb{R}^{n} \mid z_i > 0, \sum_i z_i = 1 \}$. This means that the distribution over the random variable $x$ is described by a categorical distribution with an $n$-dimensional vector of parameters $\boldsymbol{\phi}$, where $n$ is the cardinality of the sample space of $X$. Numerically, these categorical distributions can be represented as multidimensional arrays (also known as NDarrays or tensors) that contain their parameters. These categorical distributions come in two flavors: vector-valued marginal distributions (e.g. $P(x)$) usually playing the role of priors, and conditional categorical distributions (e.g. $P(y | x)$) in the form of NDarrays (matrices and tensors), playing the role of likelihoods in the generative model.  

Marginal categorical distributions are encoded in \texttt{pymdp} as simple 1-D vectors, which technically are instances of \verb+numpy.ndarrays+, the core data structure for representing multi-dimensional arrays in the Python array programming library, \verb+numpy+ \cite{harris2020array}. One can easily instantiate categorical distributions in \verb+numpy+ using calls to the array constructor, e.g. \newline\verb+ prior_over_states = np.array([0.5, 0.5])+. Conditional categorical distributions are just collections of 1-D categorical vectors, with as many 1-D vectors as there are levels of the conditioning variable. In \texttt{pymdp}, we encode these collections as matrices (2-D NDarrays) and higher-order NDarrays. For instance, we would encode some discrete conditional distribution relating two categorical variables $P(y | x)$ as a matrix of size $N \times M$, where $N$ is the number of levels of the support random variable $y$ and $M$ is the number of levels of the conditioning random variable $x$. We represent such conditional categorical distributions mathematically with the notation $P(y \mid x) = Cat(\boldsymbol{\phi}), \boldsymbol{\phi} \in \mathbb{R}^{n \times m \times p \times ...}$, where now $\boldsymbol{\phi}$ is a matrix or tensor of parameters, whose columns $\boldsymbol{\phi}_{\bullet jkl...}$ have the properties of a single categorical distribution, i.e. $\boldsymbol{\phi}_{\bullet jkl...} \in \{z \in \mathbb{R}^{m} \mid z_i > 0, \sum_i z_i = 1 \}$.

The core distributions of the generative model -- that are encoded in this way -- are the observation likelihood $P(o_\tau | s_\tau)$ (also known as the observation model or the sensory likelihood) and the transition likelihood $P(s_\tau | s_{\tau-1}, u_{\tau-1})$ (also known as the transition model or the dynamics model). For all the following descriptions, we borrow notation from the \texttt{SPM} and \texttt{DEM} standards, which are summarized in papers like \cite{friston2015active, friston2017process, smith2022step, friston2017active}. In \texttt{pymdp} notation, the observation likelihood is constructed as the \texttt{A} array. In simple generative models\footnote{See the section on \nameref{mult_factor} for the more general form.}, \texttt{A} will be a $O \times S$ matrix, where $O$ is the number of outcomes or levels of the observations $o$, and $S$ is the number of levels of hidden states $s$. The entry \texttt{A[i,j]} encodes the probability of seeing observation $i$ given state $j$. In other words, each column of this matrix \texttt{A[:,j]} stores a vector of categorical parameters that encodes the distribution $P(o_\tau | s_\tau = s_j)$. Similarly, the transition likelihood is represented by the \texttt{B} array which is a $S \times S \times U$ NDarray or tensor, where $U$ is the number of levels of the control states, and entry \texttt{B[i,j,k]} encodes the probability transitioning to state $i$ at time $t$ from state $j$ at time $t-1$, when control state or action $k$ is taken by the agent. The general structure of numerical representations of conditional distributions in \texttt{pymdp} can be expressed as follows: the first dimensions (or rows) of the matrix or NDarray represent the support of the conditional distribution, while the lagging dimensions (columns, slices, etc.) represent the random variables being conditioned on. Thus, for the observation likelihood $P(o_\tau | s_\tau)$, the first dimension of \texttt{A} represents the support of $o$ (and will have length $O$) while the second dimension represents the support of $s$ (and will have length $S$).

Beyond the \texttt{A} and \texttt{B} arrays, one can also specify an initial prior over states $P(s_1)$ -- in \texttt{pymdp} this is called the \texttt{D} vector (of length $S$) and represents the agent's beliefs about the distribution over hidden states at the first timestep of the time horizon (when $\tau = 1$).

Finally, in order to achieve goal-directed behavior under active inference, it is necessary to build a representation of some desired state or goal into the generative model. In reinforcement learning this is handled using reward functions but in active inference we instead specify a prior distribution over observations, also known as the `prior preferences' or `goal distribution' \cite{friston_reinforcement_2009}. Inference over control states is then biased by this preference distribution, leading agents to choose actions that bring them to states that (they expect) will lead to preferred observations. In \texttt{pymdp}, this is represented by the \texttt{C} array of length $O$. The default \texttt{C} array is a vector that is time-independent (the same \texttt{C} is used for all timesteps), but it is also possible to specify a time-dependent $C$ array. This can be used to represent goals that change over time or the desire to reach a specific goal in a time-dependent manner . 

After the generative model has been specified in terms of a set of likelihood and prior distributions, one can build an active inference agent in a single line using the \texttt{Agent()} constructor: e.g.\verb+ my_agent = Agent(A=A, B=B, ...)+. The \texttt{Agent()} constructor requires \texttt{A} and \texttt{B} arrays as mandatory input, while \texttt{C} and \texttt{D} vectors can be optionally included (the defaults are uniform distributions for each).

The various methods of the resulting \texttt{Agent} instance can then be used to perform active inference. 

\begin{lstlisting}[language=Python, caption=Detailed example of building and running an active inference process in \texttt{pymdp}.,label={lst:usage2}]
    import numpy as np
    
    import pymdp
    from pymdp import utils, maths
    from pymdp.agent import Agent
    
    # create a simple model with one hidden state factor, and one observation modality
    
    n_obs = 3
    n_states = 3
    
    A = utils.obj_array(1)
    A[0] = np.array([[1.0, 0.0, 0.0],
                     [0.0, 1.0, 0.0],
                     [0.0, 0.0, 1.0]])
    
    # introduce uncertainty into one of the hidden states
    inv_temperature = 0.5
    A[0][:,2] = maths.softmax(inv_temperature* A[0][:,2])
    
    # create a simple transition model with two possible actions
    
    B = utils.obj_array(1)
    B[0] = np.zeros((3, 3, 2))
    
    # first action leads to first two states with uncertainty
    B[0][:,:,0] = np.array([[0.5, 0.5, 0.5], 
                            [0.5, 0.5, 0.5],
                            [0.0, 0.0, 0.0]])
    
    # second action leads to last state with certainty
    B[0][:,:,1] = np.array([[0.0, 0.0, 0.0], 
                            [0.0, 0.0, 0.0],
                            [1.0, 1.0, 1.0]])

    # specify prior preferences (C vector)
    C = utils.obj_array_uniform([n_obs])
    
    # specify prior over hidden states (D vector)
    D = utils.obj_array(1)
    D[0] = utils.onehot(1, n_states)
    
    # instantiate your agent with a call to the `Agent()` constructor
    my_agent = Agent(A=A, B=B, C=C, D=D)
    
    # write a simple environment class, where state depends on the action probabilistically, and observation is deterministic function of the state except for state 2, where it's randomly sampled
    
    from pymdp.envs import Env
    
    # sub-class it from the base Env class
    class custom_env(Env):
        
        def __init__(self):
            self.state = 0
        
        def step(self, action):
            
            if action == 0:
                self.state = 0 if np.random.rand() > 0.5 else 1
            if action == 1:
                self.state = 2
                
            if self.state == 0:
                obs = 0
            elif self.state == 1:
                obs = 1
            elif self.state == 2:
                obs = np.random.randint(3)
            
            return obs
    
    env = custom_env()

    action = 0

    T = 10 # length of active inference loop in time
    for t in range(T):
        
        # sample an observation from the environment
        o_t = env.step(action)
        
        # do active inference
        qs = my_agent.infer_states([o_t]) # get posterior over hidden states
        my_agent.infer_policies() 
        action = my_agent.sample_action() 
        
        # convert action into int, for use with environment
        action = int(action.squeeze())
    
\end{lstlisting}

\subsection*{Specifying an environment}
For most use-cases of active inference, the agent will need to interface with some kind of environment or external world. The minimal definition of an environment is just a class or function that takes actions of the agent as input, updates the true hidden state of the environment (but does not convey this information to the agent) and returns observations generated by the updated hidden state. In the Bayesian modelling literature, this environment is also abstractly referred to as the `generative process' or `data-generating process'. What is important to note is that this generative process does not have to be identical to the generative model -- i.e. there is no requirement that an active inference agent with a POMDP generative model is operating in a world with discrete, POMDP-like dynamics \cite{baltieri2019generative,tschantz2020learning}. All that matters is that the environment accepts the agent's actions and returns observations that are discrete and are compatible with the support of the likelihood $P(o_\tau | s_\tau)$ of the agent's generative model.

\texttt{pymdp} contains a library of pre-built environments which can be imported using \texttt{from pymdp import envs}. Following the convention of OpenAI Gym \cite{brockman2016openai}, users can also write their own environment class. This class is traditionally written to have a \texttt{step()} method which takes an action from the agent as input and returns observations that will be processed by the agent at the next timestep. In many reinforcement learning and control problem contexts, the environment has its own internal state that is updated by the agent's action, and which determines (either stochastically or deterministically) the next observation.

\subsection*{Closing the action-perception loop}

The typical `active inference loop' consists of three main steps: 1) sampling an observation from the environment; 2) updating the agent's beliefs about states and policies using the observation; and 3) choosing an action, based on the agent's posterior over policies (see \nameref{sec:Appendix_B} for more details on state and policy inference). In \texttt{pymdp}, 1) is implemented by calling the environment class \texttt{env.step()}; 2) is implemented using the \texttt{pymdp} functions \texttt{agent.infer\_states()} and \texttt{agent.infer\_policies()} and 3) is implemented using the \texttt{pymdp} function \texttt{agent.sample\_action()}. Wrapping these three steps into a loop over time entails the entire active inference process; see the full example using the \texttt{Agent} class in Example \ref{lst:usage2}.

\subsection*{\label{mult_factor}Factorized representations}
Although many simple POMDPs can be constructed with simple 2-D \texttt{A} matrices and 3-D \texttt{B} arrays, most of the interesting applications of active inference require what are referred to as `factorized representations'. This requires building additional structure into the generative model, such that observations $\mathbf{o}$ are divided into separate \emph{modalities} and hidden states $\mathbf{s}$ into separate \emph{factors}. A multi-modality observation $\mathbf{o}_t$ and multi-factor hidden state $\mathbf{s}_t$ can be expressed as follows:
\begin{align}
    \mathbf{o}_t = \left\{o_t^{1}, o_t^{2},..., o_t^{M} \right\} \hspace{10mm} \mathbf{s}_t = \left\{s_t^{1}, s_t^{2}, ..., s_t^{F} \right\} \notag 
\end{align}
where here the superscript refers to the index of the $m$\textsuperscript{th} observation modality or $f$\textsuperscript{th} hidden state factor, respectively. This means that at any given time the agent receives a collection of discrete observations, where each observation within the collection belongs to a distinct `modality'. The name modality is used to emphasize the analogy to different sensory channels (e.g. vision, audition, somatosensation) in biology that relay different sorts of information. Likewise, in a factorized hidden state representation, the environment's structure is represented through several hidden state factors, that may encode distinct features of the world, each of which may have its own dimensionality, dynamics, and relationship to observations.

Importantly, with such factorized representations, the likelihood arrays \texttt{A} and \texttt{B} become more complex. In both \texttt{pymdp} and \texttt{SPM}, we encode a multi-modality \texttt{A} array as a collection of sub-arrays \texttt{A[m]}, with one for each observation modality. Each modality-specific \texttt{A} array then represents the conditional probability of observations for modality $m$, given the different configuration of hidden states, i.e., $P(o_m | \mathbf{s})$. Note that the `first' index of the larger \texttt{A} array selects a particular modality from the collection, e.g. \texttt{A\_modality = A[m]}, whereas the subsequent multi-index \emph{into the modality-specific} \texttt{A} \emph{array} selects conditional probabilities or arrays of such probabilities, e.g., \texttt{A\_modality[0, 2, 3, ...]}. Each \texttt{A[m]} thus encodes all probabilistic dependencies between the different hidden state factors hidden states and observations for the $m$\textsuperscript{th} modality: $P(o^m | \mathbf{s}) = P(o^m | s^1, s^2, ..., s^F)$. These complex conditional relationships are encoded by accordingly higher-dimensional NDarrays in \texttt{NumPy}, with the number of lagging dimensions encoding the number of hidden state factors that the observations depend on. Such factorized generative models require more involved belief updating algorithms to achieve posterior inference, usually invoking a factorized approximate posterior (e.g. a mean-field factorization), where the full posterior over all hidden state factors is factorized into a product of marginals $Q(\mathbf{s}) = \prod_{i=1}^F Q(s^i)$, where each $Q(s^i)$ is the posterior for hidden state factor $i$. Fortunately, \texttt{pymdp} easily accommodates such higher-dimensional, factorized generative models and will automatically perform message passing with respect to such generative models with an arbitrary number of observation modalities and hidden state factors. See \nameref{sec:Appendix_A} for more details on multi-factor generative models.

\subsection*{\label{demo_code}Pedagogical materials \& code}

For more example code detailing how to use \texttt{pymdp} to simulate active inference in discrete state-space environments, we refer the reader to the tutorials found in the official documentation for the repository: \href{https://pymdp-rtd.readthedocs.io}{https://pymdp-rtd.readthedocs.io/}.

\subsection*{\label{Customizability}Customizability}

\texttt{pymdp} offers a high degree of customizability in designing bespoke active inference processes, such that the methods of the \texttt{Agent} class can be called in any particular order, depending on the application, and furthermore they can be specified with various keyword arguments that entail choices of implementation details at lower levels.

For instance, if one wanted to model a purely `perceptual' task, i.e., where the agent has no ability to act, but is only concerned with hidden state estimation, then one could write an active inference loop where the \texttt{Agent} class only uses the \texttt{infer\_states()} function. This offers an advantage over the main function used to perform active inference in \texttt{SPM}, \texttt{spm\_MDP\_VB\_X.m} where customization applications are limited and in practice are implemented by modifying parts of the function by hand to suit one's needs (e.g. commenting out certain sections or adding in bespoke computations).

Moreover, by retaining a modular structure throughout the package's dependency hierarchy, \texttt{pymdp} also affords the ability to flexibly compose different low level functions. This allows users to customize and integrate their active inference loops with desired inference algorithms and policy selection routines. For instance, one could sub-class the \texttt{Agent} class and write a customized \texttt{step()} function, that combines whichever components of active inference one is interested in.  

\section*{\label{sec:other}Related software packages}

The \texttt{DEM} toolbox within \texttt{SPM} in MATLAB is the current gold-standard in active inference modelling. In particular, simulating an active inference process in \texttt{DEM} consists of defining the generative model in terms of a fixed set of matrices and vectors, and then calling the \texttt{spm\_MDP\_VB\_X.m} function to simulate a sequence of trials. \texttt{pymdp}, by contrast, provides a user-friendly and modular development experience, with core functionality split up into different libraries that separately perform the computations of active inference in a standalone fashion. Moreover, \texttt{pymdp} provides the user the ability to write an active inference process at different levels of abstraction depending on the user's level of expertise or skill with the package -- ranging from the high level \texttt{Agent} functionality, which allows the user to define and simulate an active inference agent in just a few lines of code, all the way to specifying a particular variational inference algorithm (e.g. marginal-message passing) for the agent to use during state estimation. In \texttt{SPM}, this would require setting undocumented flags or else manually editing the routines in \texttt{spm\_MDP\_VB\_X.m} to enable or disable bespoke functionality.

\texttt{pymdp} has extensive, organized documentation and illustrative examples. While the \texttt{DEM} toolbox is also replete with interesting examples that result in beautiful visualizations of simulated behavior and synthetic neural responses, the available usage information for each function remains limited to doc-strings in the source code. The closest to documentation or an instruction manual for \texttt{spm\_MDP\_VB\_X.m} is the comprehensive tutorial by Smith et al. 2021 \cite{smith2022step}, which features a series of \texttt{MATLAB} tutorial scripts that walk through the different aspects of active inference, with a focus on applications to modelling (behavioral and neurophysiological) empirical data.

A recent related, but largely non-overlapping project is \href{https://github.com/biaslab/ForneyLab.jl}{ForneyLab}, which provides a set of Julia libraries for performing approximate Bayesian inference via message passing on Forney Factor Graphs \cite{ForneyLab2019}. Notably, this package has also seen several applications in simulating active inference processes, using ForneyLab as the backend for the inference algorithms employed by an active inference agent \cite{van2019simulating, vanderbroeck2019active,ergul2020learning, van2021chance}. While ForneyLab focuses on including a rigorous set of message passing routines that can be used to simulate active inference agents, \texttt{pymdp} is specifically designed to help users quickly build agents (regardless of their underlying inference routines) and plug them into arbitrary environments to run active inference in a few easy steps.

\paragraph{Funding Statement} 
CH and IDC acknowledge support from the Office of Naval Research grant (ONR, N00014- 64019-1-2556), with IDC further acknowledging support from the European Union’s Horizon 2020 research and innovation programme under the Marie Skłodowska-Curie grant agreement (ID: 860949), the Deutsche Forschungsgemeinschaft (DFG, German Research Foundation) under Germany’s Excellence Strategy-EXC 2117- 422037984, and the Max Planck Society. KF is supported by funding for the Wellcome Centre for Human Neuroimaging (Ref: 205103/Z/16/Z) and the Canada-UK Artificial Intelligence Initiative (Ref: ES/T01279X/1). CH, DD, and BK acknowledge the support of a grant from the John Templeton Foundation (61780). The opinions expressed in this publication are those of the author(s) and do not necessarily reflect the views of the John Templeton Foundation.

\paragraph{Acknowledgements} 

The authors would like to thank Dimitrije Markovic, Arun Niranjan, Sivan Altinakar, Mahault Albarracin, Alex Kiefer, Magnus Koudahl, Ryan Smith, Casper Hesp, and Maxwell Ramstead for discussions and feedback that contributed to development of \texttt{pymdp}. We would also like to thank Thomas Parr for pointing out a technical error in an earlier version of the paper. Finally, we are grateful to the many users of \texttt{pymdp} whose feedback and usage of the package have contributed to its continued improvement and development.

\printbibliography[title={References}]

\clearpage

\appendix
\setcounter{figure}{0}
\setcounter{table}{0}
\renewcommand\thefigure{\thesection.\arabic{figure}}    
\renewcommand\thetable{\thesection.\arabic{table}}  

\section*{Appendix A: Factorized Generative Models}
\label{sec:Appendix_A}

In this appendix, we explain observation and state factorization by using an in-depth example. Let's imagine a scenario where you have to infer two simultaneous states of the world, given some sensory data. The two facts you need to estimate are 1) what time of day it is (morning, midday, or evening), and 2) whether it rained recently (yes or no). We can represent this in a generative model as an environment characterized by two discrete random variables or hidden state factors. The first variable or factor we can call the \emph{time-of-day} state, which has three levels: Morning, Midday, or Evening; the second factor we can call the \emph{did-it-rain} state, which has two levels: Rained and Did-not-rain. We use the following notation to denote these hidden state factors and their respective levels:
\begin{align}
    \mathbf{s} &= \{s^{\textrm{time-of-day}}, s^{\textrm{did-it-rain}} \} \notag \\
    s^{\textrm{time-of-day}} \in \{Mo,  Mid, &Eve\} \hspace{10mm} s^{\textrm{did-it-rain}} \in \{Rain,  NoRain\} \notag
\end{align}
where we assign each state level a cardinal index, e.g. $Mo = 0$, $Mid = 1$, $Eve = 2$, and $Rain = 0, NoRain = 1$. Let's now augment our simple hidden state representation with observations, which a hypothetical agent would use to infer both the time of day and whether it rained recently, i.e. to obtain a posterior distribution over $\mathbf{s}$: $P(\mathbf{s} \mid \mathbf{o})$. Our observations will also be factorized, into two different \emph{modalities} or information channels. Each of these modalities is also a discrete-valued random variable. Let's imagine our two modalities are: 1) the ambient light level (dark, cloudy, or sunny) and 2) humidity (dry or humid). We denote the observations as follows:
\begin{align}
    \mathbf{o} &= \{o^{\textrm{light}}, o^{\textrm{hum}} \} \notag \\
    o^{\textrm{light}} \in \{Drk, Cld, &Sun\} \hspace{10mm} o^{\textrm{humid}} \in \{Dry, Hmd\} \notag 
\end{align}

Having specified a factorized representation of both states and observations, we can now consider how observations lend evidence for or against different states-of-affairs in the environment. For example, if you notice it's dark outside (i.e., $o^{\textrm{light}} = Drk$), that provides evidence to suggest that it's night time, rather than being morning or midday. At the same time, you might also notice that the air is humid through your humidity modality, i.e. $o^{\textrm{hum}} = Hmd$. We can imagine that the humidity observation provides no evidence for the time of day it is, but it may suggest that it rained recently. 

These probabilistic relationships between the observation modalities and the hidden state factors, which are used to perform inference, are encoded in the observation likelihood $P(o_m | \mathbf{s})$, represented in \texttt{pymdp} as a modality-specific sub-array \texttt{A[m]}. Recall that each likelihood array encodes the conditional dependencies between each setting of the hidden states $\mathbf{s}$ and the observations within modality $m$, i.e. \texttt{A[m]} = $P(o_m | s_1, s_2, ..., s_F)$. Therefore the dimensionality of a given \texttt{A[m]} array is $O_m \times S_1 \times ... \times S_F$, where $O_m$ is the dimensionality of modality $m$ and $S_f$ is the dimensionality of factor $f$. Following our simple example of inferring the time of day and whether it just rained, the likelihood for the first modality would be a 3D array that represents the likelihood distribution $P(o^{\textrm{light}} | s^{\textrm{time-of-day}}, s^{\textrm{did-it-rain}})$. Specifically, entry \texttt{A[i,j,k]} encodes the probability of observing $o^{\textrm{light}} = $ level $i$, given $s^{\textrm{time-of-day}} = $ level $j$ and $s^{\textrm{did-it-rain}} = $ level $k$. The second observation modality accordingly has its own likelihood NDarray, encoding the likelihood distribution $P(o^{\textrm{hum}} | s^{\textrm{time-of-day}}, s^{\textrm{did-it-rain}})$.

These higher-dimensional likelihood arrays enable complex, conjunctive relationships to be encoded in the generative model. For instance, we might imagine that the $o^{\textrm{light}}$ observation depends on both the time of day \emph{and} whether it just rained. For instance, all else being equal we might expect the ambient lighting to be sunny, if the time of day is midday. However, if it just rained then the probability of the ambient lighting being dark or cloudy might be higher, even if the time of day is midday. This statement already requires a nonlinear, conjunctive relationship between $s^{\textrm{time-of-day}}$ and $s^{\textrm{did-it-rain}}$. For example the probability distribution over the 3 levels of $o^{\textrm{light}}$, given $s^{\textrm{time-of-day}}$ is $Mo$ and $s^{\textrm{did-it-rain}}$ is $Rain$ would be encoded by the corresponding likelihood: $P(o^{\textrm{light}} | s^{\textrm{time-of-day}} = Mo, s^{\textrm{did-it-rain}} = Rain )$. This would then be easily encoded in the corresponding `slice' of the high-dimensional \texttt{A[light]} array: \texttt{A[light][:,Mo,Rain]}, where \texttt{light} is the index of the \texttt{A} array that encodes the $o^{\textrm{light}}$ likelihood, and \texttt{Mo} and \texttt{Rain} are the cardinal indices for those factor-specific state levels.

In the same way that the observation likelihood for each modality $m$ is represented using a single likelihood NDarray \texttt{A[m]}, the transition likelihood for each hidden state factor $f$ is represented using a single likelihood NDarray \texttt{B[f]}, where the size of the $f$\textsuperscript{th} transition array is of size $S_f \times S_f \times U_f$. So analogously to having a collection of \texttt{A[m]} arrays, one for each observation modality, we also have a collection of \texttt{B[f]} arrays, one for each state factor. Two important things to note are that: 1) constructing B matrices with this $S_f \times S_f \times U_f$ shape assumes that hidden state factors cannot influence each other dynamically, i.e. the next state within a factor $s^f_t$ only depends on the past state \emph{for that factor} $s^f_{t-1}$ and control state $u^{f}_{t-1}$, and 2) that control states are factorized just like hidden states, such that for each hidden state factor there is a corresponding \textit{control factor} $c^{f}$ whose dimensionality is equal to the number of control state levels or actions that can be taken upon hidden state factor $f$. It is of course allowable to have \emph{uncontrollable} hidden state factors - in which case we simply set the dimensionality of the corresponding control factor to 1, i.e. $U_f = 1$. We often refer to these as `trivial' control factors, since they don't actually encode any kind of control.

The factorized representation described above offers several advantages. First of all, if a large hidden state space can be factorized into a collection of one-dimensional representations, then the memory cost of storing the relevant probability distributions over hidden states can be greatly finessed (e.g. $P(s_t | s_{t-1})$, $P(s)$, etc.). For example, if you can represent the identity of an object and its location independently, without having to enumerate all the combinations of both its identity and its location together, then the amount of memory used to store the factorized representation will be linear in the dimensionality of the two hidden state factors, whereas the `enumerated' representation will be polynomial. For example, if location is a 1000-dimensional vector, and identity is a 1000-dimensional vector, then storing two 1000-dimensional vectors is considerably cheaper than storing a single $1000 \times 1000$-dimensional vector.

Another advantage is the degree of interpretability and model transparency that factorized representations afford; a particular factorization is ideally explicitly designed, such that  hidden state factors are directly mapped to intuitive features of the environment whose relationships are easy to reason about. If a multi-factor model of, for example, 3 hidden state factors (e.g. the location, identity, and time of some event) were fully enumerated into a single, 1-dimensional hidden state, then each level of the single hidden state would correspond to a unique combination of ``what'', ``where'' and ``when''. When it comes to encoding probabilistic relationships in the generative model (e.g. the observation and transition models), it becomes harder to visualize and reason about the relationships between such high-dimensional state combinations. Thus factorization also proves a useful tool when designing generative models based on prior domain or task knowledge.

Interestingly, when one optimizes the factorial structure of a generative model, using marginal likelihood or variational free energy, the best factorisation maximizes marginal likelihood (a.k.a., model evidence) by minimizing complexity: namely, the degrees of freedom used to provide an accurate account of observations. This is an important aspect of active inference; namely, that to provide the best account of observations -- that precludes overfitting and ensures generalization -- the (mean-field) factorisation should be as simple as possible but no simpler.

Finally, inference also may take advantage of the factorized structure of the generative model. In doing so, inference is not only more memory-efficient, but the belief-updating algorithms have features like functional specialization \cite{mishkin1983object, zeki1991direct} and local message-passing that have been linked to features of computation in the brain \cite{jardri2017experimental, leptourgos2020circular}. It has been argued that this affords factorized generative models a higher degree of biological plausibility \cite{parr2019neuronal, parr2020modules}.

\section*{Appendix B: Modules and Theory}
\label{sec:Appendix_B}

\subsection*{\texttt{inference.py}}
\label{sec:inference_module}

In this section we provide an overview of the \texttt{inference} module of \texttt{pymdp} and then briefly rehearse the mathematics of variational inference, both generally and as it is used in \texttt{pymdp}.

The \texttt{inference.py} file contains functions for performing variational inference about hidden states in discrete categorical generative models. Functions within this module are called by the \texttt{infer\_states()} method of \texttt{Agent}. The core functions of this module are:
\begin{itemize}
    \item \texttt{update\_posterior\_states(obs, A, prior=None, **kwargs)}: This function computes the variational (categorical) posterior over hidden states at the current timestep $Q(\mathbf{s}_t)$. This function by default calls the standard or `vanilla' inference algorithm offered by \texttt{pymdp}, which estimates the marginal posteriors for each hidden state factor at the current timestep $Q(s^i_t)$ using fixed-point iteration. This function can be generically applied as the inference step in any discrete POMDP model, as all it requires are some observations \texttt{obs}, a likelihood array \texttt{A} and optionally a prior over hidden states \texttt{prior}, which will have the same structure as the resulting posterior. The additional arguments \texttt{**kwargs} contain parameters that will be passed to the \texttt{run\_vanilla\_fpi()} function in \texttt{algos/fpi.py} (see the \href{https://pymdp-rtd.readthedocs.io}{documentation} for more details).
    
    \item \texttt{update\_posterior\_states\_full(A,B,prev\_obs,policies, prev\_actions=None,\newline prior=None, policy\_sep\_prior=True, **kwargs}): This function computes the variational (categorical) posterior over hidden states under all policies: $Q(\tilde{\mathbf{s}}|\pi)$. The notation $\tilde{\mathbf{s}}$ represents a trajectory of hidden states over time. This is inspired by the `full construct' active inference process as implemented in \texttt{spm\_MDP\_VB\_X.m} in \texttt{DEM}, where the full posterior over both hidden states and policies is computed: $Q(\mathbf{s}, \mathbf{\pi})$.  This function itself calls the \texttt{run\_mmp.py} function within the \texttt{algos} library, which estimates the marginal posteriors for hidden state factor $i$, at time point $\tau$ under policy $j$: $Q(s^i_{\tau} \mid \pi_j)$, using marginal message passing \cite{parr2019neuronal} (for more details, see the \texttt{algos.py} summary below). This function calls \texttt{run\_mmp.py} once per policy, estimating the policy-conditioned posterior over all timepoints of the horizon and for all hidden state factors. The number of timepoints over which inference occurs is not necessarily identical to the total time horizon of the simulation: rather this time horizon is a function of the number of previous observations (\texttt{len(prev\_obs)}) and the temporal depth of the policy under consideration (\texttt{len(policies[j])}). Thus the hidden state beliefs indexed by $\tau$ refer to a finite time horizon that is relative to the current timestep $t$, where this horizon $[t - H_0, t + H_1]$ has a `lookback' length $H_0$ and a `planning horizon' $H_1$. As arguments, this function requires the observation (\texttt{A}) and transition (\texttt{B}) likelihoods of a generative model, a list of previous (including current) observations (\texttt{prev\_obs}), a list of policies (\texttt{policies}), an optional list of actions taken up until the current timepoint (\texttt{prev\_actions}) an optional prior over hidden states at the start of the time horizon (\texttt{prior}), and a keyword argument \texttt{policy\_sep\_prior}, which determines whether the prior is itself conditioned on policies $P(\mathbf{s}_0 | \pi)$ vs. unconditioned on policies $P(\mathbf{s}_0)$. The additional \texttt{**kwargs} contain parameters that will be passed to the \texttt{run\_mmp()} function in \texttt{algos/mmp.py}. For more details on \texttt{run\_mmp()}, see the \href{https://pymdp-rtd.readthedocs.io}{documentation}.
    
\end{itemize}

Specific usage examples--in addition to descriptions of other specialized functions in \texttt{inference.py}--are more extensively covered in the \href{https://pymdp-rtd.readthedocs.io}{official documentation}.

\subsubsection*{Bayesian and Variational Inference}

A central task in statistics is to perform inference, which can be mathematically represented as computing posterior distributions of one variable given another from a joint distribution. For example, suppose we are given some observation $o$, and we then want to infer the likely state $s$ underlying that observation. Critical to achieving this task is the possession of a generative model, or joint distribution, $P(o,s)$ that tells us how observations and hidden states are related. We can formulate the problem of inferring $s$ as finding a posterior distribution over the states, given the observation: $P(s | o)$. We can compute this posterior distribution using our generative model and Bayes Rule:
\begin{align}
    P(s | o) = \frac{P(o,s)}{P(o)} \notag
\end{align}

The generative model $P(o,s)$ is often factorized into a likelihood $P(o|s)$ and a prior $P(s)$, while $P(o)$ is known as the \emph{marginal likelihood} or \emph{model evidence} and can be computed by solving the integral $P(o) = \int P(o,s)ds$. This expresses the idea that the marginal probability of observations $o$ in $P(o)$ is the sum (or integral) over all the ways that that $o$ depends on $s$, for all possible settings of $s$.

While Bayes rule provides a simple formula for relating the posterior distribution to the generative model, explicitly computing this distribution can often be difficult in practice due to the computational expense involved in performing the integral over all states necessary to compute the marginal likelihood $P(o)$. However, a number of approximate Bayesian inference methods have been developed which circumvent this computational difficulty at the expense of only returning approximately correct posterior distributions.

Variational inference \cite{beal2003variational,wainwright2008graphical} is a widely used and well understood approach for performing approximate Bayesian inference. The central idea in variational inference is that instead of directly computing the posterior, we instead optimize the parameters $\theta$ of an arbitrary distribution $Q(s; \theta)$ so as to minimize the divergence between this distribution and the true posterior. This arbitrary distribution is often named the \textit{approximate} posterior, because in the course of minimizing the divergence, the arbitrary distribution becomes an approximation to the true posterior, i.e. $Q(s;\theta) \approx P(s | o)$. In this way, variational inference converts a challenging inference problem (involving computing intractable integrals) into a relatively straightforward optimization problem, for which many powerful algorithms exist in the optimization literature.

Ideally, variational inference would directly minimize the Kullback-Leibler divergence between the approximate and true posteriors:
\begin{align}
    \theta^* = \underset{\theta}{\operatorname{argmin}} \, \, \operatorname{D}_{KL}[Q(s; \theta)||P(s|o)] \notag
\end{align}

In the present form,, this objective is also intractable since it depends on the true posterior $P(s|o)$, whose approximation is our goal. However, by supplementing the KL divergence with the log marginal likelihood, which does not depend upon $Q(s;\theta)$, we can convert the above KL divergence into an \emph{upper bound} on the log marginal likelihood, called \emph{variational free energy} (VFE) $\mathcal{F}$. Crucially, this can be rearranged into a computable form:

\begin{align}
    \theta^* &=  \underset{\theta}{\operatorname{argmin}} \, \, \mathcal{F} 
    \notag \\
    \mathcal{F} &= \operatorname{D}_{KL}[Q(s; \theta)\parallel P(s|o)] - \ln P(o) \notag\\
    &= \mathbb{E}_{Q}[\ln Q(s; \theta) - \ln P(s,o)] \notag
\end{align}

Thus, by minimizing $\mathcal{F}$, we minimize the divergence between the approximate and true posterior, thus forcing the approximate posterior to more closely resemble the true one. Moreover, if this optimization finds the exact solution, such that $KL[Q(s; \theta)||P(s|o)] = 0$ then value of $\mathcal{F} = -\ln P(o)$ provides the marginal likelihood ($P(o) \propto e^{-\mathcal{F}}$) which can then be used for model selection and structure learning. More generally, $\mathcal{F}$ is known as an evidence bound, because the KL divergence can never be less zero \cite{mackay1992bayesian, penny2004modelling}.

\subsubsection*{Variational Inference in \texttt{pymdp}}

Active inference agents in \texttt{pymdp} perform inference over both hidden states $s$ and policies $\pi$ in a POMDP generative model. In this appendix, we only consider inference over states, while inference over policies is treated in the following section \nameref{sec:control}.

Recall the POMDP generative model is a joint distribution over observations, states, policies, and parameters. For the purposes of hidden state inference, we will condition the whole generative model on some fixed policy $\pi$, so that we can re-write it as follows:

\begin{align}
    P(o_{1:T}, s_{1:T},\phi \mid \pi) = P(\phi)P(s_1)\prod_{\tau = 2}^{T} P(s_{\tau} | s_{\tau-1}, \pi,\phi)\prod_{\tau = 1}^{T} P(o_\tau | s_\tau,\phi) \label{eq:vfe_definition}
\end{align}

For an active inference agent equipped with this generative model, instantaneous inference consists in optimizing an approximation to the posterior over the current hidden state: $P(s_{\tau} | s_{\backslash\tau}, o_{[1:\tau]})$ given the past and future states $s_{\backslash\tau} = \{s_{[1:\tau-1]}, s_{[\tau+1 : T]} \}$, and the observations collected up until the current timepoint $o_{[1:\tau]}$. Mathematically, this inference can be described as minimizing the following free energy over trajectories with respect to the variational parameters $\theta$:
\begin{align}
    \theta^{*} &= \underset{\theta}{\operatorname{argmin}}\hspace{1mm} \mathcal{F}_{1:T} \notag \\ 
    \mathcal{F}_{1:T} &= \mathbb{E}_{Q}\left[\ln Q(\mathbf{s}_{1:T};\theta) - \ln P(o_{1:T}, s_{1:T}; \phi \mid \pi)\right] \notag
\end{align}

Thus the goal of hidden state inference is the optimization of variational parameters $\theta$ which parameterise the approximate posterior over hidden states: $Q(\mathbf{s}_{1:T}; \theta)$. In the case of our discrete POMDP generative model, the variational parameters $\theta$ are the sufficient statistics of categorical distributions. Fortunately, these parameters are easy to interpret, since they are identical to the probabilities of sampling each outcome level in the distribution's support: i.e. $Q(s; \theta) = \mathbf{Cat}(\theta)$. For example, the variational parameters of some categorical distribution $Q(s) = \begin{bmatrix} 0.1 & 0.4 & 0.5 \end{bmatrix}$ would simply be $\theta = \begin{bmatrix} 0.1 & 0.4 & 0.5 \end{bmatrix}$. In the equations to follow, we therefore exclude the variational parameters $\theta$ when writing the variational posterior, referring to it hereafter as simply $Q(\mathbf{s})$. Below we describe the variational inference methods currently offered by \texttt{pymdp} as of this document's writing (November 2021).

The \texttt{run\_vanilla\_FPI} function (within \texttt{algos/fpi.py}) implements an inference algorithm known as \emph{fixed-point iteration} to optimize the posterior over hidden states $Q(\mathbf{s}_{\tau})$ at a given timestep $\tau$. Central to this algorithm is the assumption of a factorized structure to the variational posterior, such that the posterior at time $\tau = i$ is independent of the posterior at any other timestep $\tau = j$, where $j \neq i$. In addition to this temporal factorization, we further assume the posterior at a given timestep $Q(\mathbf{s}_\tau)$ is factorized across different hidden state factors: i.e. $Q(s^{f}_{\tau})$ is independent of $Q(s^{f'}_{\tau})$ (see the section \nameref{mult_factor} for more on multi-factor hidden states). This factorization is also known as a \emph{mean-field} approximation in the statistics and physics literatures and can be expressed as follows: 
\begin{align}
    Q(\mathbf{s}_{[1:T]}) = \overset{T}{\underset{\tau = 1}{\prod}} Q(\mathbf{s}_\tau) \hspace{10mm} Q(\mathbf{s}_{\tau}) = \overset{F}{\underset{f = 1}{\prod}} Q(s^f_\tau) \notag
\end{align}

Given this factorization, the full free energy over trajectories now also factorizes into a sum of free energies across time, which can be minimized independently of each other. Thus, for a given time $\tau$, we can write the time-dependent free energy\footnote{For the remainder of the section we remove the hyperparameters $\phi$ from the generative model and approximate posterior, but inference over these are treated in the section on \nameref{sec:learning}}:
\begin{align}
    \mathcal{F}_\tau &= \mathbb{E}_{Q}\left[\ln Q(\mathbf{s}_\tau) - \ln P(\mathbf{o}_\tau | \mathbf{s}_\tau)P(\mathbf{s}_\tau | \mathbf{s}_{\tau-1}, \mathbf{u}_{\tau -1})\right]
\end{align}
where we now use the bold notation $\mathbf{s}$ and $\mathbf{o}$ to express potentially multi-factor (or -modal) hidden states (or observations) in the generative model, in the same way that the variational posterior is factorized. Inference proceeds by optimizing $Q(\mathbf{s}_\tau)$ in order to minimize the timestep-specific free energy $\mathcal{F}_{\tau}$.

Importantly, we can solve for the variational posterior analytically for a given timestep $\tau$ and factor $f$ by setting the derivative of the free energy $\mathcal{F}_{\tau}$ to $0$ and solving for $Q(s^{f}_\tau)$. We express this partial derivative as $\frac{\partial \mathcal{F}_\tau}{\partial q^{f}}$, and can express it as follows:
\begin{align}
    \frac{\partial \mathcal{F}_\tau}{\partial q^{f}} &= \frac{\partial}{\partial q^{f}} \left[\sum Q(\mathbf{s}_\tau) \left( \ln Q(\mathbf{s}_\tau)  -  \ln P(\mathbf{o}_{\tau}, \mathbf{s}_{\tau}) \right)\right]= 0 \notag \\
    &= \ln Q(s^{f}_\tau) + \mathbf{1} -  \mathbb{E}_{q^{i \backslash f}}\left[\ln P(\mathbf{o}_\tau | \mathbf{s}_\tau)- \ln \left(\mathbb{E}_{P(\mathbf{s}_{\tau-1}, \mathbf{u}_{\tau-1})}[P(\mathbf{s}_\tau | \mathbf{s}_{\tau-1}, \mathbf{u}_{\tau -1})]\right)\right] \ = 0 \notag \\
    &\implies \ln Q(s^{f}_\tau) = \mathbb{E}_{q^{i \backslash f}}\left[\ln P(\mathbf{o}_\tau | \mathbf{s}_\tau)\right] + \ln \left(\mathbb{E}_{P(s^{f}_{\tau-1}, u^{f}_{\tau-1})}[P(s^{f}_\tau | s^{f}_{\tau-1}, u^{f}_{\tau -1})]\right) - \mathbf{1} \notag \\
    &\implies Q^{*}(s^{f}_\tau) = \sigma\left(\mathbb{E}_{q^{i \backslash f}}\left[\ln P(\mathbf{o}_\tau | \mathbf{s}_\tau)\right] + \ln \left(\mathbb{E}_{P(s^{f}_{\tau-1}, u^{f}_{\tau-1})}[P(s^{f}_\tau | s^{f}_{\tau-1}, u^{f}_{\tau -1})]\right)\right) \label{eq:fixed_point_update}
\end{align}
where the expectation $\mathbb{E}_{q^{i \backslash f}}$ denotes an expectation with respect to all posterior marginals $Q(s^i_\tau)$ besides the marginal $Q(s^f_\tau)$ currently being optimized, and  $\sigma(x) = \frac{e^{x}}{\sum_x e^{x}}$ is a normalized exponential or softmax function. The update equation for each marginal posterior offers an intuitive Bayesian interpretation, where the belief about the current state is the product of an observation likelihood term $P(\mathbf{o}_\tau | \mathbf{s}_\tau)$ and a `prior' term, $\mathbb{E}_{P(\mathbf{s}_{\tau-1}, \mathbf{u}_{\tau-1})}[P(\mathbf{s}_\tau | \mathbf{s}_{\tau-1}, \mathbf{u}_{\tau -1})]$, where the prior is dynamically determined by the previous state, previous action, and the transition likelihood. Note that in practice we set the prior at any given timestep equal to the posterior optimized at the previous timestep, i.e. $P(\mathbf{s}_{\tau - 1}, \mathbf{u}_{\tau - 1}) \equiv Q^{*}(\mathbf{s}_{\tau-1}, \mathbf{u}_{\tau-1}) = Q^{*}(\mathbf{s}_{\tau-1})Q^{*}(\mathbf{u}_{\tau-1})$, where $Q^{*}(\mathbf{u}_{\tau-1})$ is a Dirac delta function over the action actually taken (the agent has perfect knowledge of the action it just took). This means that the prior term in the last line of Equation \eqref{eq:fixed_point_update} can be rewritten as $\mathbb{E}_{P(s^{f}_{\tau-1}, u^{f}_{\tau-1})}[P(s^{f}_\tau | s^{f}_{\tau-1}, u^{f}_{\tau -1})] = \mathbb{E}_{Q(s^f_{\tau-1}, u^f_{\tau-1})}[P(s^{f}_\tau | s^{f}_{\tau-1}, u^{f}_{\tau -1})]$. In the \texttt{run\_vanilla\_fpi.py} function of \texttt{pymdp}, the fixed point equation is solved iteratively for each marginal posterior $Q(s^f_\tau)$, using the latest solution for the other marginals $Q(s^{i \backslash f}_\tau)$ to compute the expected log-likelihood term for the marginal $f$ currently being updated: $\mathbb{E}_{q^{i \backslash f}}\left[\ln P(\mathbf{o}_\tau | \mathbf{s}_\tau)\right]$. Note that the prior $P(s^{f}_\tau | s^{f}_{\tau-1}, u^{f}_{\tau -1})P(s^{f}_{\tau-1})$ only depends on the marginal $f$ currently being updated, because in \texttt{pymdp} the transition likelihoods are assumed to be independent across hidden states, i.e. hidden states from factor $i$ do not determine the dynamics of hidden states of another factor $j$ (see \nameref{sec:Appendix_A} for details). Given enough iterations,\footnote{The default number of iterations for \texttt{run\_vanilla\_fpi()} function is \texttt{num\_iter=10} but in many generative models with precise likelihood arrays (i.e. low entropy rows/columns), convergence is often achieved in many fewer iterations. This is further controlled by a tolerance parameter that tracks the change in the free energy across iterations.} the fixed point equations converge to a unique solution of the variational posterior $Q^{*}(\mathbf{s}_\tau)$.

In \texttt{pymdp}, the approximate posterior $Q(\mathbf{s})$ is represented as a collection of 1-D \texttt{NumPy} arrays (e.g. \texttt{qs}), where individual elements of the collection (e.g. \texttt{qs[f]}) store the marginal posterior for a particular hidden state factor. The likelihood distributions are represented by \texttt{A} and \texttt{B} arrays. If we take the limiting case of a generative model and variational posterior with a single hidden state factor, the update equation for the variational posterior at a given timestep using fixed point iteration reduces to a single line of \texttt{NumPy} code:
\begin{verbatim}
    qs_current = softmax(np.log(A[o,:])+ np.log(B[:,:,u_last].dot(qs_last)))
\end{verbatim}
where utility functions like \texttt{softmax} are available from the \texttt{utils.py} and \texttt{maths.py} modules of \texttt{pymdp}.
In the default initialization of the \texttt{Agent} class, the \texttt{infer\_states()} method will call the \texttt{update\_posterior\_states()} function of the \texttt{inference} module; this in turns calls upon fixed point iteration (via \texttt{run\_vanilla\_fpi()}) to update the variational posterior over hidden states. Therefore all that is required for inference at a given timestep is to provide some observation \texttt{obs} to \texttt{infer\_states()}. The prior over hidden states is automatically updated within the \texttt{Agent} class, where \texttt{prior} will either A) equal the initial belief about hidden states (the \texttt{D} vector) in the case that $\tau = 1$ ; or B) the dot product of the transition likelihood conditioned on the last action \texttt{B[:,:,u\_last]}, and the posterior at the last timestep \texttt{qs\_last} in the case that $\tau > 1$: \texttt{prior=B[:,:,u\_last].dot(qs\_last)}.

Another, more complex inference algorithm known as \emph{marginal message passing} (MMP) is also implemented in the \texttt{run\_mmp()} function, also found within the \texttt{algos} module. Marginal message passing makes weaker assumptions about the factorization of the variational posterior, and incorporates the computational advantages of two well-known message passing algorithms: belief propagation \cite{yedidia2000generalized} and variational message passing \cite{winn2005variational}. In practice, using marginal-message passing instead of standard fixed-point iteration enables more accurate inference due to its less restrictive assumptions as to the form of the variational posterior, at the expense of additional computational cost. For the purpose of brevity and since this algorithm has been discussed in detail elsewhere (specifically, see Appendix C of \cite{friston2017process} as well a comprehensive treatment in \cite{parr2019neuronal}), we will not describe the mathematics behind marginal message passing here. It is worth noting that for beginning users, a standard active inference simulation will not require marginal-message passing to achieve the desired behavior; state inference achieved with instantaneous fixed-point iteration often suffices for practitioners interested in simulating a target behavior. However, cognitive neuroscientists are often interested in modelling neuronal responses based on estimated inferential dynamics. In this case, more sophisticated schemes like \texttt{run\_mmp()} may be required, where actual dynamics of belief updating might be used as a forward model of hypothesized electrophysiological processes (e.g. local field potentials or spiking activity). Finally, we also mention that in order to achieve identical behavior to active inference agents simulated using \texttt{spm\_MDP\_VB\_X.m}, it is necessary to use \texttt{run\_mmp()}. 

In practice, a desired inference algorithm can be specified by passing the name of the algorithm into the \texttt{Agent()} constructor, e.g. \texttt{my\_agent = Agent(...,inference\_algo=`MMP')}. For more detailed instructions on how to initialize an \texttt{Agent} with different customization options, please see the \href{https://pymdp-rtd.readthedocs.io}{documentation}.

\subsection*{\texttt{control.py}}
\label{sec:control}

The core functions implementing policy inference and action selection (i.e. control) in \texttt{pymdp} can be found in the \texttt{control.py} file. As with \texttt{inference}, these functions are called by methods of \texttt{Agent} like \texttt{infer\_policies()} and \texttt{sample\_action()}, but can also be directly imported from the \texttt{control} library and used for custom applications. Below we briefly summarize the core functions of the \texttt{control} module:

\begin{itemize}
    \item \texttt{update\_posterior\_policies(qs,  }  \texttt{A,  } \texttt{B,  } \texttt{C,  } \texttt{policies,  } \texttt{use\_utility=True,\newline use\_states\_info\_gain=True, use\_param\_info\_gain=False, pA=None, pB=None,\newline E=None, gamma=16.0)}: This function computes the posterior over policies $Q(\pi)$ using an initial posterior belief about hidden states at the current timestep \texttt{qs}. Specifically, it computes the expected free energy of each policy (discussed further in the next section) by summing the expected free energies over a future path in the case of multi-timestep or `temporally-deep' policies. This function first loops over all policies, computes the expected states and observations under each policy, and then sums the expected free energies calculated from those predicted future states and observations. The expected free energy for each policy is combined with its prior probability under the generative model $P(\pi$) (in \texttt{pymdp} represented by the \texttt{E} vector) and softmaxed to determine the posterior over policies $Q(\pi)$ (in \texttt{pymdp} represented by \texttt{q\_pi}). Optional Boolean parameters like \texttt{use\_utility} and \texttt{use\_states\_info\_gain} can be turned on and off to selectively enable (disable) computation of components of the expected free energy (see the section on \nameref{sec:EFE} for information on these different expected free energy terms).
    \item \texttt{update\_posterior\_policies\_full(qs\_seq\_pi, A,B,C, policies,use\_utility= \newline True, use\_states\_info\_gain=True, use\_param\_info\_gain=False, prior=None,\newline pA=None,  pB=None, } \texttt{ F=None, E=None, gamma=16.0)}: This function computes the posterior over policies $Q(\pi)$ using a posterior belief over hidden states over multiple timesteps under all policies. This version differs from the standard function, \texttt{update\_posterior\_policies()}, in that the expected hidden states over future timepoints, under different policies, have already been computed in the input, the posterior beliefs \texttt{qs\_seq\_pi}. This function for policy inference should thus be used in tandem with the `advanced' inference schemes (like marginal message passing) where posterior beliefs over multiple timesteps, under all policies, are computed during the inference step. As a consequence, this function only computes the expected \emph{observations} under each policy for all future timesteps, and then uses the expected states (already part of the inputs) and expected observations under all policies to calculate the expected free energy for each policy. This is integrated with prior belief about policies \texttt{E} and the variational free energy of policies \texttt{F} (see the section on \nameref{subsec: control_policy_inference} for more information on the variational free energy of policies) to finally determine the posterior over policies $Q(\pi)$, often represented in \texttt{pymdp} as \texttt{q\_pi}. This function's remaining arguments (e.g. \texttt{use\_utility}) are identical to how they are used in the standard \texttt{update\_posterior\_policies()} function.
    \item \texttt{get\_expected\_states(qs, B, policy)}: This function computes a posterior distribution over future states given a current state distribution (\texttt{qs}), a transition model (\texttt{B}) and a policy (\texttt{policy}). Specifically, this function projects the current beliefs about hidden states forward in time by iteratively taking the inner product of \texttt{qs} with the action-conditioned \texttt{B} matrix, where the actions are those entailed by the \texttt{policy}.
    \item \texttt{get\_expected\_obs(qs\_pi, A)}: This function computes the observations expected under a (policy-conditioned) hidden state distribution \texttt{qs\_pi}. In the case of a sequence of hidden states over time, \texttt{qs\_pi} will be a list of hidden states distributions with one element per timestep e.g. \texttt{qs\_pi[t]}. This function only requires an expected state distribution \texttt{qs\_pi} and an observation model \texttt{A}.
    \item \texttt{calc\_expected\_utility(qo\_pi, C)}: This function computes the extrinsic value or utility part of the expected free energy using the prior preferences or `goal distribution' encoded by the \texttt{C} vector. The \texttt{C} is encoded in terms of relative log probabilities and thus need not be a proper probability distribution.
    \item \texttt{calc\_states\_info\_gain(A, qs\_pi)}: This function computes intrinsic value or information gain  part of the expected free energy, in particular the information gain or epistemic value about hidden states $s$.
    \item \texttt{calc\_pA\_info\_gain(pA, qo\_pi, qs\_pi)}: This function computes the information gain about the Dirichlet prior parameters over the observation model (\texttt{A}) , also known as the `novelty' term of the expected free energy \cite{friston2017active}. It requires a Dirichlet prior over the observation model \texttt{pA}, an expected observation distribution \texttt{qo\_pi} and an expected state distribution \texttt{qs\_pi}. It is recommended to include this information gain term in the expected free energy calculation, when also simultaneously performing \texttt{A} array learning (i.e. inference over Dirichlet hyperparameters), since it leads to the agent exploring regions which lead to the largest updates of the parameters of \texttt{A} .
    \item \texttt{calc\_pB\_info\_gain(pB, qs\_pi, qs\_prev, policy)}: This function computes the information gain about the Dirichlet prior parameters over the transition model (\texttt{B}), also known as the `novelty' term of the expected free energy. It requires a Dirichlet prior over the transition model \texttt{pB}, an expected state distribution under a policy \texttt{qs\_pi}, an initial state distribution \texttt{qs\_prev}, and a policy \texttt{policy}. It is recommended to include this information gain term in the expected free energy calculation, when also simultaneously performing \texttt{B} array learning (i.e. inference over Dirichlet hyperparameters).
    \item \texttt{construct\_policies(num\_states, num\_controls=None, }\texttt{policy\_len=1,  \newline} \texttt{control\_fac\_idx=None)}: This is a utility function which builds an array of policies by combinatorially enumerating them from a set of actions and a time horizon. It can be used to construct a full set of policies based on the time horizon and action space of the environment, if the policy set is not explicitly stated by the user.
    \item \texttt{sample\_action(q\_pi, policies, num\_controls, \newline action\_selection="deterministic", alpha=16.0)}: This function samples an action, given the posterior distribution over policies and a desired sampling scheme. In particular, this function computes the posterior over control states $u$ by marginalising the posterior over policies with respect to each control state, i.e. $Q(u_t) = \sum_\pi P(u_t \mid \pi) Q(\pi)$, where $P(u_t \mid \pi)$ is the mapping between policies and control states. To obtain an action, the most probable action is either A) selected deterministically (\texttt{action\_selection="deterministic"}) as the most probable control state or B) an action is sampled from the control posterior (\texttt{action\_selection="stochastic"}), using a Boltzmann distribution with inverse temperature given by \texttt{alpha}.
    
\end{itemize}

\subsubsection*{Control in Active Inference}

Policy inference consists in computing the `goodness' or `quality' of each policy, given the ability to compute the expected consequences of each policy and the agent's goals. In active inference this is done by using a quasi-utility function known in the literature as the \emph{Expected Free Energy} (EFE) (often denoted $\mathbf{G}$). Under active inference, agents are equipped with a particular prior over policies $P(\pi)$ that assumes policies are inversely proportional to the free energy expected under their pursuit, i.e.:

\begin{align}
    P(\pi) = \sigma(-\mathbf{G})
\end{align} 

Equipped with this policy prior in the generative model, active inference agents perform policy inference by optimizing $Q(\pi)$, the variational posterior over policies. As we shall see in the section \nameref{subsec: control_policy_inference}, computing $Q(\pi)$ entails computing the expected free energy of each policy (the contribution from the prior) as well as the variational free energy of policies (analogous to the `evidence' for each policy).

\paragraph*{The Expected Free Energy}\label{sec:EFE}

The expected free energy is the crucial component that determines the behavior of active inference agents. The EFE is designed to be similar to the VFE of standard variational inference but with two major modifications to enable its use as an objective which, when minimized, will perform goal seeking behavior rather than simply inference. Firstly, since the EFE ranks future performance, where future observations are not known, it contains an \emph{expectation} over future observations. Secondly, as there needs to be a way to integrate the notion of goals or rewards into the inference procedure, the EFE alters the generative model of the agent to be `biased' in such a way that it predicts the agent reaches rewarding or \emph{a priori} preferred states \cite{parr2019generalised}. Thus, performing inference to maximize the likelihood of visiting these rewarding states naturally leads to policies that help the agent achieve its goals. Moreover, an additional benefit is that minimizing the EFE also entails an exploratory, inherently uncertainty-reducing component to behavior. This endows behavior with an additional `epistemic drive' which aids in computing the optimal long-term policies \cite{friston2015active}. For in-depth discussion of the nature of the EFE and the exploratory drive it induces please see \cite{friston2015active,friston2017process,millidge2021whence,millidge2021understanding}.

The expected free energy is a function of observations, states, and policies, and is defined mathematically as:
\begin{align}
    \mathbf{G}(o_{1:T}, s_{1:T},\pi) = \mathbb{E}_{Q}[\ln Q(s_{1:T},\pi) - \ln \tilde{P}(o_{1:T}, s_{1:T},\pi)] \label{eq:EFE_definition}
\end{align}
where $\tilde{P}$ represents a generative model `biased' towards the preferences of the agent. We can write this predictive generative model at a single timestep, under a given policy, as $\tilde{P}(o_\tau, s_\tau |\pi) = P(s_\tau | o_\tau, \pi) \tilde{P}(o_\tau)$, where $\tilde{P}(o_\tau)$ represents a `predictive prior' over observations, represented in \texttt{pymdp} with the \texttt{C} array. Given the factorization of the approximate posterior $Q(s, \pi)$ over time, the EFE for a single policy and timestep can also be defined as follows:
\begin{align}
    \mathbf{G}_{\tau}(\pi) &= \mathbb{E}_{Q(o_\tau, s_\tau | \pi)}[\ln Q(s_\tau | \pi) - \ln \tilde{P}(o_\tau, s_{\tau} |\pi)] \notag\\
    &= \mathbb{E}_{Q(o_\tau, s_\tau | \pi)}\left[ \ln Q(s_\tau | \pi) - \ln \tilde{P}(o_\tau, s_\tau | \pi) + \underbrace{\ln Q(s_\tau | o_\tau, \pi) - \ln Q(s_\tau | o_\tau, \pi)}_{=0}\right] \notag\\ 
    &= \mathbb{E}_{Q(o_\tau, s_\tau | \pi)}\left[Q(s_\tau | \pi) - Q(s_\tau | o_\tau, \pi) - \ln \tilde{P}(o_\tau)\right] + \underbrace{\mathbb{E}_{Q(o_\tau |\pi)}\left[\operatorname{D}_{KL}[Q(s_\tau | o_\tau) \parallel P(s_\tau | o_\tau, \pi)]\right]}_{\text{Expected approximation error} \geq 0} \notag\\
    &\geq -\underbrace{\mathbb{E}_{Q(o_\tau|\pi)}\left[\operatorname{D}_{KL}[Q(s_\tau|o_\tau, \pi)\parallel Q(s_\tau | \pi)]\right]}_{\text{Epistemic Value}} - \underbrace{\mathbb{E}_{Q(o_\tau|\pi)}[\ln\tilde{P}(o_\tau)]}_{\text{Utility}} \label{eq:EFE_decomposition}
\end{align}
where the first term, the epistemic value \cite{friston2015active}, encourages the pursuit of policies expected to yield high information gain about hidden states, expressed here as the divergence between the states predicted under a policy, with and without conditioning on observations. The second term represents the degree to which expected outcomes under a policy will align with prior preferences over observations. Since the prior over policies is \emph{inversely} proportional to the expected free energy, policies will thus be more likely if they visit states that resolve uncertainty (maximize epistemic value) and satisfy prior preferences (maximize utility). The epistemic value terms give active inference agents a degree of superior exploration capacity compared to standard reinforcement learning agents. In \texttt{pymdp}, the EFE is computed using exactly this decomposition into epistemic value and utility, where the expected approximation error (penultimate line of Equation \eqref{eq:EFE_decomposition}) is implicitly assumed to be 0, so the bound becomes equality.  The utility term is computed by the function \texttt{calc\_expected\_utility()} while the epistemic value term (also known as the information gain) is computed by the function \texttt{calc\_states\_info\_gain()}. Both of these functions are found within \texttt{control.py}. The computation of the utility term is particularly straightforward for categorical distributions, since it reduces to the dot product of the expected observations under a policy $Q(o_\tau | \pi)$ with the log of the prior preferences or `goal vector' $\tilde{P}(o_\tau)$, i.e. the \texttt{C} array.

\paragraph*{\label{control_param_info_gain}Parameter Information Gain}
In the case where the agent also maintains a variational posterior over parameters $Q(\phi)$, the timestep- and policy-dependent EFE has an augmented form, since it needs to account for the expected information gain over both hidden states and parameters \cite{friston2017active}:
\begin{align}
    \mathbf{G}_{\tau}(\pi) &= \mathbb{E}_{Q(o_\tau, s_\tau | \pi)}[\ln Q(s_\tau, \phi|\pi) - \ln\tilde{P}(o_\tau, s_\tau, \phi |\pi)] \notag \\
    &\approx \mathbb{E}_{Q(o_\tau, s_\tau |\pi)}[Q(s_\tau|\pi) - Q(s_\tau, o_\tau|\pi) + Q(\phi | \pi) - Q(\phi, o_\tau|\pi) - \ln\tilde{P}(o_\tau)] \notag \\
    &= -\underbrace{\mathbb{E}_{Q(o_\tau|\pi)}\left[\operatorname{D}_{KL}[Q(s_\tau|o_\tau, \pi)\parallel Q(s_\tau | \pi)]\right]}_{\text{(State) Epistemic Value}} \notag \\
    &\hspace{5mm}-\underbrace{\mathbb{E}_{Q(o_\tau|\pi)}\left[\operatorname{D}_{KL}[Q(\phi|o_\tau, \pi)\parallel Q(\phi | \pi)]\right]}_{\text{(Parameter) Epistemic Value}} - \underbrace{\mathbb{E}_{Q(o_\tau|\pi)}[\ln\tilde{P}(o_\tau)]}_{\text{Utility}}
    \label{eq:EFE_with_phi}
\end{align}

So now the EFE is supplemented with an additional epistemic value, the so called `parameter' epistemic value or `parameter information gain'. This additional term arises when the approximate posterior includes variational beliefs about model hyperparameters: $Q(s_\tau, \phi) = Q(s_\tau)Q(\phi)$. The optimization of the posterior over model parameters $Q(\phi)$ is handled in the next section on the \nameref{sec:learning} module. This presence of this term in the expected free energy mediates what's also been referred to as `active learning' or `model exploration', i.e. the drive to resolve uncertainty about the parameters of one's generative model \cite{schwartenbeck2019computational}. 

In the discrete state space case implemented in \texttt{pymdp}, this parameter epistemic value is computed with respect to the Dirichlet parameters (conjugate priors over categorical distributions) that parameterise the prior and approximate posterior over the likelihoods and priors over the generative model, i.e. the \texttt{A}, \texttt{B}, \texttt{C} and \texttt{D} arrays. This is implemented as of the time of writing (December 2021) for information gain about the  parameters of the \texttt{A} array and \texttt{B} array, parameterised respectively by the Dirichlet conjugate priors \text{pA} and \texttt{pB}. The relevant functions for computing these information gains are  \texttt{calc\_pA\_info\_gain()}. and \texttt{calc\_pB\_info\_gain()}. 

\paragraph*{\label{subsec: control_policy_inference}Policy Inference}

Given the definition of the expected free energy in Equations \eqref{eq:EFE_definition} and \eqref{eq:EFE_decomposition}, we now are equipped to describe posterior inference over policies, i.e., how to obtain $Q(\pi)$. 

We begin by expanding the variational free energy $\mathcal{F}$ as defined in Equation \eqref{eq:vfe_definition}, dropping parameters $\phi$ for simplicity:

\begin{align}
    \mathcal{F}_{1:T} &= \mathbb{E}_{Q(s_{1:T}, \pi})\left[\ln Q(s_{1:T}, \pi) - \ln P(o_{1:T}, s_{1:T}, \pi)\right] \notag \\
    &= \mathbb{E}_{Q(s_{1:T}, \pi)}[\ln Q(\pi) + \sum_{\tau = 1}^{T}\ln Q(s_\tau |\pi) - \ln P(\pi) - \ln P(o_{1:T}, s_{1:T} |\pi)] \notag \\
    &= \operatorname{D}_{KL}[Q(\pi)\parallel P(\pi)] + \mathbb{E}_{Q(\pi)}\left[F(\pi)\right] \label{eq:full_VFE_with_policies}
\end{align}

where the variational free energy of a particular policy $F(\pi)$ is defined as follows:

\begin{align}
    F(\pi) &= -\mathbb{E}_{Q(s_{1:T} |\pi)}[\ln P(o_{1:T}, s_{1:T} | \pi) - \mathbf{H}[Q(s_{1:T}|\pi)]
\end{align}

The optimal posterior that minimizes the full variational free energy $\mathcal{F}$ is found by taking the derivative of $\mathcal{F}$ with respect to $Q(\pi)$ and setting this gradient to $0$, yielding the following free-energy-minimizing solution for $Q(\pi)$:

\begin{align}
    Q^{*}(\pi) &= \underset{Q(\pi)}{\operatorname{argmin}}\hspace{2mm}\mathcal{F} = \sigma(\ln P(\pi) - F(\pi)) 
\end{align}

where the prior over policies $P(\pi)$ is the softmax of the negative expected free energy $\sigma(-\mathbf{G}(\pi))$. Note that in the case of "temporally deep" or multi-timestep policies, the expected free energy of a given policy $\mathbf{G}(\pi)$ is the sum of the timestep-specific expected free energies:

\begin{align}
    \mathbf{G}(\pi) = \sum_{\tau} \mathbf{G}_{\tau}(\pi)
\end{align}

In \texttt{pymdp} and the \texttt{DEM} toolbox of MATLAB, one has the option of augmenting the prior over policies with a `baseline policy' or `habit vector' $P(\pi_0)$, also referred to as the \texttt{E} vector. This means the full expression for the optimal posterior can be written as (expanding $\ln P(\pi)$ as $\ln P(\pi_0) - \mathbf{G}(\pi)$):

\begin{align}
    Q^{*}(\pi) = \sigma( -\mathbf{G}(\pi) + \ln P(\pi_0) - F(\pi))
\end{align}

This means the inferred policy distribution combines influences from the expected free energy of each policy ($\mathbf{G}_{\pi}$), a baseline prior probability assigned to each policy ($\ln P(\pi_0)$) and the variational free energy of each policy ($F(\pi)$). Numerically, policy inference is achieved by computing the expected and variational free energies of each policy and then combining them with the policy prior (the \texttt{E} vector) before softmaxing them. The expected free energy is computed per policy as the integral of the timestep-specific expected free energies, as shown in Equation \eqref{eq:EFE_decomposition}. This is achieved by computing the `posterior predictive densities' expected under each policy: $Q(o_{t:T}, s_{t:T}|\pi)$ and using those densities to compute and add together the epistemic value and utility for each policy. These posterior predictive densities are simply the posterior beliefs at the current timestep $t$ `multiplied through' the transition and observation models (in code: the \texttt{A} and \texttt{B} arrays) over the temporal horizon of the policy. By doing this iteratively across policies, the $\mathbf{G}$ vector ends up storing a `cost' for each policy, which is then integrated with the policy prior and variational free energy of each policy to determine the posterior probability of each policy, stored in $Q(\pi)$. This boils down to the following single line of \texttt{NumPy} code:

\begin{verbatim}
    q_pi = softmax(-G + np.log(E) - F)
\end{verbatim}

In \texttt{pymdp}, the functions \texttt{update\_posterior\_policies()} and \texttt{update\_posterior\_policies\newline\_full()} of the \texttt{control} module perform the calculations needed for policy inference, and themselves are called by the \texttt{infer\_policies} method of \texttt{Agent}.

\subsection*{\texttt{learning.py}}
\label{sec:learning}
In this section we will summarize the functions in the \texttt{learning} module and then derive the update equations for updating model parameters of the likelihood and prior distributions that comprise POMDP generative models.

The functions used to implement inference over model parameters can be found in the \texttt{learning.py} file. These functions are called by methods of \texttt{Agent} like \texttt{update\_A()}, \texttt{update\_B()}, and \texttt{update\_D()}. We survey the most important functions of the \texttt{learning} module below:

\begin{itemize}
    \item \texttt{update\_obs\_likelihood\_dirichlet(pA, A, obs, qs, lr=1.0, modalities="all")}:\newline This function computes the posterior Dirichlet parameters $Q(A)$ over the \texttt{A} array or observation model $P(o_\tau|s_\tau, A)$. As input arguments this function requires the current Dirichlet prior \texttt{pA} over the parameters of the \texttt{A} array, the current value of the categorical \texttt{A} array (which is also the expected value of the Dirichlet prior \texttt{pA}), an observation \texttt{obs}, the current posterior beliefs about hidden states \texttt{qs}, a learning rate \texttt{lr} and a list of which observation modalities to update, \texttt{modalities}. The default setting is to update the \texttt{A} arrays associated with all observation modalities (\texttt{modalities = "all"}), but this extra argument allows one to only update specific sub-arrays of a larger multi-modality \texttt{A} array. For example, \texttt{modalities = [0, 1]} would only update sub-arrays \texttt{A[0]} and \texttt{A[1]}. The learning rate parameter scales the size of the update to the posterior over the \texttt{A} array.
    \item \texttt{update\_state\_likelihood\_dirichlet(pB, B, actions, qs, qs\_prev, lr=1.0, \newline factors="all")}: This function computes the posterior Dirichlet parameters $Q(B)$ over the \texttt{B} array or transition model $P(s_{\tau}|s_{\tau-1}, u_{\tau-1}, B)$. As input arguments this function requires the current Dirichlet prior \texttt{pB} over the parameters of the \texttt{B} array, the current value of the categorical \texttt{B} array, the posterior beliefs about hidden states at the current timestep \texttt{qs}, the posterior beliefs about hidden states at the previous timestep \texttt{qs\_prev}, a learning rate \texttt{lr} and a list of which hidden state factors to update, \texttt{factors}. The default setting is to update the \texttt{B} arrays associated with all hidden state factors (\texttt{factors = "all"}), but this extra argument allows you to only update specific sub-arrays of a larger multi-factor \texttt{B} array. For example, \texttt{factors = [0, 1]} would only update sub-arrays \texttt{B[0]} and \texttt{B[1]}. The learning rate parameter scales the size of the update to the posterior over the \texttt{B} array.
    \item \texttt{update\_state\_prior\_dirichlet(pD, qs, lr=1.0, factors="all")}: This function computes the posterior Dirichlet parameters $Q(D)$ over the \texttt{D} array or prior over initial hidden states $P(s_0| D)$. As input this function requires the current Dirichlet prior \texttt{pD} over the parameters of the \texttt{D} array, the posterior beliefs about hidden states at the current timestep \texttt{qs}, a learning rate \texttt{lr} and a list of which hidden state factors to update, \texttt{factors}. The default setting is to update the \texttt{D} vectors associated with all hidden state factors (\texttt{factors = "all"}), but this extra argument allows you to only update specific sub-vectors of a larger multi-factor \texttt{D} array. For example, \texttt{factors = [0, 1]} would only update sub-arrays \texttt{D[0]} and \texttt{D[1]}. The learning rate parameter scales the size of the update to the posterior over the \texttt{D} array.
\end{itemize}

\paragraph*{\label{subsec:learning_maths}Inference of POMDP model parameters}
Under active inference, learning is cast as inference about model parameters, and in the context of neuroscience is often analogized to slower-scale changes to inter-neuronal synaptic weights (e.g. Hebbian learning) \cite{da2020active}. Parameter inference is referred to as `learning' because it is often assumed to occur on a fundamentally slower timescale than hidden state and policy inference \cite{friston2016active}. However, the update equations for model parameters follow the exact same principles as hidden state inference - namely, we optimize a variational posterior over model parameters $Q(\phi)$ by minimizing the variational free energy $\mathcal{F}$. 

For the POMDP generative models used in \texttt{pymdp}, learning manifests as posterior inference over hyperparameters of the (categorical) likelihood and priors of the generative model. We use Dirichlet distributions as conjugate priors for the categorical distributions\footnote{a prior is called \textit{conjugate} to a likelihood when the resulting posterior is the same distribution family as the prior}, meaning that the hyperparameters $\phi$ become the parameters of Dirichlet distributions. This choice of parameterization results in remarkably simple and biologically-plausible updates for the posteriors over these parameters, wherein `fire-together-wire-together'-like Hebbian increments are used to learn the parameters as a function of observations. Below we derive the update rule for Dirichlet hyperparameters over the \texttt{A}, \texttt{B}, and \texttt{D} arrays.

To begin, we augment the POMDP generative model in \eqref{eq:POMDP_gen_model} with the parameters of the likelihood and prior categorical distributions and Dirichlet priors over each of them. In order to do this, we divide the hyperparameters $\phi$ in the into subsets that correspond to the categorical and Dirichlet parameters over the \texttt{A}, \texttt{B}, and \texttt{D} arrays:

\begin{align}
    \phi &= \{A, a, B, b, D, d\} \notag \\
    P(o_\tau | s_\tau, A) &= \mathbf{Cat}(A) \notag \\ 
    P(A) &= \prod_{j}P(A_{\bullet j}), \hspace{3mm} P(A_{\bullet j}) = Dir(a_{\bullet j}) \notag \\ 
    P(s_\tau | s_{\tau-1},u_{\tau-1}, B) &= \mathbf{Cat}(B) \notag \\ 
    P(B) &= \prod_{j}\prod_{u}P(B_{\bullet ju}), \hspace{3mm} P(B_{\bullet ju}) = Dir(b_{\bullet ju}) \notag \\ 
    P(s_1| D) &= \mathbf{Cat}(D) \notag \\ 
    P(D) &= Dir(d) \label{eq:dirichlet_defs}
\end{align}

where the notation $X_{\bullet j}$ denotes the $j\textsuperscript{th}$ column of a matrix $X$. Under this parameterisation, $A$, $B$, and $D$ are arrays of categorical parameters (i.e. probabilities) that `fill out' the entries of the \texttt{A}, \texttt{B}, and \texttt{D} arrays respectively. The Dirichlet parameters $\mathbf{a}$, $\mathbf{b}$, and $\mathbf{d}$ are similarly the parameters of Dirichlet priors over these categorical distributions, and have identical dimensionality to the distributions they parameterise. The Dirichlet parameters are constrained to be positive real numbers ($a, b, d \in \mathbb{R}_{>0}$) that score the prior probability of each entry of the categorical distribution they parameterize. Dirichlet values, like the parameters of other common conjugate prior distributions, can be interpreted as `pseudo-counts' measuring how often a particular outcome level is expected \emph{a priori} (e.g. the prior probability assigned to a particular state-observation coincidence in the case of the $A$ distribution). Note that for notational convenience we assume the generative model is not factorized into multiple hidden state factors and observation modalities, but for generality one could add in additional indices to capture multiple hidden state factors and observation modalities. For instance, the most general form of a (potentially multi-modality and multi-factor) observation model would be:

\begin{align}
    P(\mathbf{o}_\tau | \mathbf{s}_\tau, \mathbf{A}) = \{\mathbf{Cat}(A^1), \mathbf{Cat}(A^2), ..., \mathbf{Cat}(A^M) \} \notag \\
    P(A^m) = \prod_{j,k,...}P(A^m_{\bullet jk...}), \hspace{3mm} P(A^m_{\bullet jk...}) = Dir(a^m_{\bullet jk...}) \notag
\end{align}

Given the introduction of the new Dirichlet priors in Equation \eqref{eq:dirichlet_defs}, we can now write down the augmented generative model, where the hyperparameters $\phi$ have been split into individual priors over $A$, $B$, and $D$:

\begin{align}
    P(o_{[1:T]}, s_{[1:T]}, \pi, A, B, D) = P(A)P(B)P(D)P(s_1|D)\prod_{\tau=2}^{T}P(s_\tau|s_{\tau-1},B)\prod_{\tau=1}^{T}P(o_\tau|s_\tau, A)  \label{eq:POMDP_gen_model_v2}
\end{align}

Given the new generative model with Dirichlet priors, we can now formulate learning as approximate inference about these parameters, i.e. optimizing variational posteriors over the likelihood and prior parameters. We begin by expanding our expression of the variational posterior to include beliefs over the values of the $A$, $B$, and $D$ distributions:

\begin{align}
Q(s_{[1:T]}, \pi, A, B, D)
&= Q(A)Q(B)Q(D)Q(\pi)\prod_{\tau = 1}^{T}Q(s_\tau |\pi)
\notag \\
\textrm{where} \hspace{5mm} Q(A) &= \prod_{j}Q(A_{\bullet j}), \hspace{3mm}Q(A_{\bullet j}) = Dir(\mathbf{a}_{\bullet j}) \notag \\
Q(B) &= \prod_{j}\prod_{u}Q(B_{\bullet ju}), \hspace{3mm}Q(B_{\bullet ju}) = Dir(\mathbf{b}_{\bullet ju}) \notag \\
Q(D) &= Dir(\mathbf{d})\notag
\end{align}

where now the variational parameters $\mathbf{a}$, $\mathbf{b}$, and $\mathbf{d}$
are Dirichlet parameters of the approximate posteriors $Q(A)$, $Q(B)$, and $Q(D)$, respectively. Performing inference with respect to $A$, $B$, and $D$ thus amounts to optimizing the variational Dirichlet parameters in order to minimize free energy. This is what is meant by `learning' in active inference.

We will now step through the update rules for each of the Dirichlet posteriors over the $A$, $B$, and $D$ distributions. We begin by writing down the full variational free energy:

\begin{align}
    \mathcal{F}_{1:T} &= \mathbb{E}_{Q(s_{1:T}, A, B, D, \pi)}\left[\ln Q(s_{1:T}, A, B, D, \pi) - \ln P(o_{1:T}, s_{1:T}, \pi, A, B, D)\right] \notag \\
    &= \mathbb{E}_{Q(s_{1:T}, A, B, D, \pi)}[\ln Q(A) - \ln P(A) + \ln Q(B) - \ln P(B) \notag\\ &\hspace{10mm}+ \ln Q(\pi) - \ln P(\pi) + \ln Q(D) - \ln P(s_1) \notag\\&\hspace{10mm}+ \sum_{\tau=1}^{T} \ln Q(s_\tau | \pi) - \sum_{\tau = 2}^{T}\ln P(s_\tau | s_{\tau - 1}, \pi, B) \notag\\&\hspace{10mm}- \sum_{\tau = 1}^{T}\ln P(o_\tau | s_{\tau}, A)] \label{eq:full_VFE_learning}
\end{align}

\paragraph*{\label{subsec_learning_A} Learning the observation model} 
We begin with the update rule for the Dirichlet parameters over $Q(A)$, i.e. updating the parameters of the observation model or \texttt{A} array. We can first isolate the components of the free energy that depend on $Q(A)$ since we're interested in the gradients of the free energy with respect to $\mathbf{a}$, the parameters of $Q(A)$:

\begin{align}
    F_{1:T} &= \operatorname{D}_{KL}[Q(A)\parallel P(A)] - \sum_{\tau = 1}^{T}\mathbb{E}_{Q(s_\tau, \pi)Q(A)}[\ln P(o_\tau | s_{\tau}, A)] + ... \label{eq:VFE_depend_A}
\end{align}

We can expand the KL divergence between $Q(A)$ and $P(A)$ as follows, using the definition of the KL divergence between Dirichlet distributions and the independence of the Dirichlet matrices across different columns (i.e. hidden state levels):

\begin{align}
    \operatorname{D}_{KL}[Q(A)\parallel P(A)] &= \sum_{j} \operatorname{D}_{KL}[Dir(\mathbf{a}_{\bullet j})\parallel Dir(a_{\bullet j})] \notag \\
    &= \sum_{j}\left( \ln \Gamma(\mathbf{a}_{0j}) - \sum_{i}\ln \Gamma(\mathbf{a}_{ij}) - \ln \Gamma(a_{0j}) + \sum_{i}\ln \Gamma(a_{ij}) \right) \notag \\
    &\hspace{2mm}+(\mathbf{a} - a)\mathbb{E}_{Q(A)}[\ln P(o_\tau | s_\tau, A)] \label{eq:kldiv_dir_A}
\end{align}

where the terms $\mathbf{a}_{0}$ and $a_{0}$ are matrices whose entries store the column-wise sums of the Dirichlet parameters of $\mathbf{a}$ and $a$, respectively, i.e. $\mathbf{a}_{0j} = \sum_{i}\mathbf{a}_{ij}$.  We can then combine the KL divergence with the remaining term in the free energy that depends on $Q(A)$ and take the gradients of the free energy with respect to $\mathbb{E}_{Q(A)}[\ln P(o_\tau | s_\tau, A)]$, which we hereafter refer to as $\boldsymbol{\ln}\mathbf{A}$:

\begin{align}
    \mathcal{F}_{1:T} &= \sum_{j}\left( \ln \Gamma(\mathbf{a}_{0j}) - \sum_{i}\ln \Gamma(\mathbf{a}_{ij}) - \ln \Gamma(a_{0j}) + \sum_{i}\ln \Gamma(a_{ij}) \right) \notag \\
    &\hspace{2mm}+(\mathbf{a} - a)\mathbb{E}_{Q(A)}[\ln P(o_\tau | s_\tau, A)] - \sum_{\tau = 1}^{T}\mathbb{E}_{Q(s_\tau, \pi)Q(A)}[\ln P(o_\tau | s_{\tau}, A)] + ...\notag \\
    \implies \frac{\partial \mathcal{F}}{\partial \boldsymbol{\ln}\mathbf{A}} &= \mathbf{a} - a - \sum_{\tau = 1}^{T} o_\tau \otimes \mathbf{s_\tau} \notag \\
    \textrm{where}\hspace{2mm} \mathbf{s_\tau} &= \sum_\pi Q(s_\tau |\pi) Q(\pi) \label{eq:dFdlnA}
\end{align}

where $\otimes$ denotes the outer product and $\mathbf{s}_\tau$ is also known as the Bayesian model average of hidden states, where the average is taken with respect to the posterior over policies, $Q(\pi)$. This move is what allows one to convert from the full posterior over hidden states and policies $Q(s_{1:T},\pi)$ to a hidden state representation that is not conditioned on policies.\footnote{The Bayesian model average can be computed using the function \texttt{average\_states\_over\_policies()} in the \texttt{inference} module.} If we set the gradient $\frac{\partial \mathcal{F}}{\partial \boldsymbol{\ln}\mathbf{A}}$ equal to 0 and solve for $\mathbf{a}$, then we obtain the fixed-point solution for the variational posterior:

\begin{align}
    \mathbf{a}^{*} = a + \sum_{\tau = 1}^{T} o_\tau \otimes \mathbf{s_\tau} \label{eq:A_matrix_update}
\end{align}

Note that the use of the gradient with respect to $\boldsymbol{\ln}\mathbf{A}$ instead of $\mathbf{a}$ directly is sufficient for deriving the update rule \cite{lance_lnA}. This can be seen by rewriting $\boldsymbol{\ln}\mathbf{A}$ as $\psi(\mathbf{a}) - \psi(\mathbf{a}_0)$, where $\psi$ is the component-wise digamma function \cite{friston2016active}. The digamma function is monotonically increasing in $\mathbf{a}$, meaning $\frac{\partial \mathcal{F}}{\partial \boldsymbol{\ln}\mathbf{A}}$ and $\frac{\partial \mathcal{F}}{\partial \mathbf{a}}$ have the same minima. This learning rule is implemented by the function \texttt{update\_obs\_likelihood\_dirichlet()} in the \texttt{learning} module, which itself is wrapped by the \texttt{update\_A()} method of \texttt{Agent}.

\paragraph*{\label{subsec_learning_B} Learning the transition model} 
The updates for the Dirichlet posterior $Q(B)$ are derived similarly to those for $Q(A)$, starting from the full expression for the variational free energy, isolating those terms that only depend on $\mathbf{b}$:

\begin{align}
    \mathcal{F}_{1:T} &= \operatorname{D}_{KL}[Q(B)\parallel P(B)] - \sum_{\tau = 2}^{T}\mathbb{E}_{Q(s_\tau, s_{\tau-1}, \pi)Q(B)}[\ln P(s_\tau | s_{\tau-1}, \pi, B)] + ... \notag \\
    &= \operatorname{D}_{KL}[Q(B)\parallel P(B)] - \sum_{\tau = 2}^{T}\mathbb{E}_{Q(\pi)}\left[Q(s_\tau|\pi)^{T}\mathbb{E}_{Q(B)}[\ln P(s_{\tau} | s_{\tau-1}, u_{\tau}, B)]Q(s_{\tau-1}|\pi)\right]\label{eq:VFE_depend_B}
\end{align}

where $u_{\tau}$ is the action expected at time $\tau$ under $Q(\pi)$. If we take the gradients of $\mathcal{F}_{1:T}$ with respect to $\boldsymbol{\ln} \mathbf{B} = \mathbb{E}_{Q(B)}[\ln P(s_{\tau} | s_{\tau-1}, u_{\tau}, B)]$ and solve for $\frac{\partial \mathcal{F}}{\partial \boldsymbol{\ln}\mathbf{B}} = 0$, then we recover the variational solution for the posterior parameters $\mathbf{b}$:

\begin{align}
    \mathbf{b}_{u}* = b_{u} + \sum_{\tau = 2}^{T}\sum_{\pi}Q(u_\tau|\pi)Q(\pi)\left(Q(s_\tau|\pi) \otimes Q(s_{\tau-1}|\pi)\right) \label{eq:B_matrix_update}
\end{align}

This update can be expressed intuitively as follows: a given timestep $\tau$ the B matrix is updated using the outer product of the beliefs about states at $\tau$ and the beliefs about states at $\tau-1$. These updates are done in an action-conditioned sense, such that the update only applies to the $u\textsuperscript{th}$ `slice' of the $\mathbf{b}$ Dirichlet parameters, depending on the action(s) expected at time $\tau$ under $Q(\pi)$. In \texttt{pymdp}, we assume that the actions at past timesteps are known with certainty, meaning that the term $Q(u_\tau |\pi)Q(\pi)$ reduces to a delta function over the action actually taken, and the update for $\mathbf{b}$ only happens to one slice $\mathbf{b}_u$ at a time. The relevant function for implementing learning $Q(B)$ is \texttt{update\_transition\_likelihood\_dirichlet()} in \texttt{learning} and the \texttt{update\_B()} method of \texttt{Agent}.

\paragraph*{\label{subsec_learning_D} Learning the state prior}
Finally, we can write down the updates for the Dirichlet posterior over initial hidden states $Q(D)$ using the same formalism as used for $Q(A)$ and $Q(B)$. First, we find the terms of the free energy that depend on $\mathbf{d}$:

\begin{align}
    \mathcal{F}_{1:T} &= \operatorname{D}_{KL}[Q(D)\parallel P(D)] - \mathbb{E}_{Q(\pi)}[Q(s_1|\pi)]^{T}\mathbb{E}_{Q(D)}[\ln P(s_1|D)]+...\label{eq:VFE_depend_D}
\end{align}

Taking the gradients of $\mathcal{F}$ with respect to $\boldsymbol{\ln}\mathbf{D} = \mathbb{E}_{Q(D)}[\ln P(s_1|D)]$ and setting the gradient to $0$ yields the following fixed-form solution for $\mathbf{d}$:

\begin{align}
    \mathbf{d}^{*} &= d + \mathbb{E}_{Q(\pi)}[Q(s_1|\pi)]\label{eq:D_vector_update}
\end{align}

The relevant function for implementing learning $Q(D)$ is \texttt{update\_state\_prior\_dirichlet()} in \texttt{learning} and the \texttt{update\_D()} method of \texttt{Agent}.

For more complete versions of each of these derivations, we refer the reader to Section 8 and Appendix A.1 of Da Costa et al. 2021 \cite{da2020active}.

\end{document}